\documentclass[letterpaper]{article} % DO NOT CHANGE THIS
\usepackage{aaai23}  % DO NOT CHANGE THIS
\usepackage{times}  % DO NOT CHANGE THIS
\usepackage{helvet}  % DO NOT CHANGE THIS
\usepackage{courier}  % DO NOT CHANGE THIS
\usepackage[hyphens]{url}  % DO NOT CHANGE THIS
\usepackage{graphicx} % DO NOT CHANGE THIS
\usepackage{natbib}  % DO NOT CHANGE THIS AND DO NOT ADD ANY OPTIONS TO IT
\usepackage{caption} % DO NOT CHANGE THIS AND DO NOT ADD ANY OPTIONS TO IT
\usepackage{algorithm}
\usepackage{algorithmic}

\usepackage{newfloat}
\usepackage{listings}
\DeclareCaptionStyle{ruled}{labelfont=normalfont,labelsep=colon,strut=off} % DO NOT CHANGE THIS
\floatstyle{ruled}
\newfloat{listing}{tb}{lst}{}
\floatname{listing}{Listing}
\title{Policy-Adaptive Estimator Selection for Off-Policy Evaluation}
\author{
    Takuma Udagawa\textsuperscript{\rm 1}, 
    Haruka Kiyohara\textsuperscript{\rm 2}\thanks{This work was done at Hanjuku-Kaso Co., Ltd.},
    Yusuke Narita\textsuperscript{\rm 3},
    Yuta Saito\textsuperscript{\rm 4},
    Kei Tateno\textsuperscript{\rm 1}
}
\affiliations{
    \textsuperscript{\rm 1}Sony Group Corporation,
    \textsuperscript{\rm 2}Tokyo Institute of Technology,
    \textsuperscript{\rm 3}Yale University,
    \textsuperscript{\rm 4}Cornell University \\
    Takuma.Udagawa@sony.com,
    kiyohara.h.aa@m.titech.ac.jp,
    yusuke.narita@yale.edu, \\
    ys552@cornell.edu,
    Kei.Tateno@sony.com
}

\usepackage{algorithm}
\usepackage{algorithmic}
\usepackage{bbding}
\usepackage{enumitem}
\usepackage{amsmath}
\usepackage{amsfonts}
\usepackage{mathtools}
\usepackage{setspace}
\usepackage{threeparttable, booktabs, makecell, caption}
\usepackage{siunitx}

\renewcommand{\algorithmicrequire}{\textbf{Input:}}
\renewcommand{\algorithmicensure}{\textbf{Output:}}

\usepackage{multicol,multirow}
\usepackage{bbding}
\usepackage{comment}
\usepackage{bm}
\usepackage{color}
\definecolor{dkgreen}{rgb}{0,0.6,0}
\definecolor{customgray}{rgb}{0.25,0.25,0.25}
\definecolor{customred}{rgb}{0.8,0.05,0.05}
\definecolor{customblue}{rgb}{0.05,0.05,0.8}

\newcommand{\emphasize}[1]{\textbf{\textit{{#1}}}}

\DeclareMathOperator*{\argmin}{arg\,min}
\DeclareMathOperator*{\argmax}{arg\,max}

\newcommand{\mE}{\mathbb{E}}
\newcommand{\mV}{\mathbb{V}}
\newcommand{\calD}{\mathcal{D}}
\newcommand{\calX}{\mathcal{X}}
\newcommand{\calA}{\mathcal{A}}

\newcommand{\calS}{\mathcal{S}}

\newcommand{\dm}{\hat{V}_{\mathrm{DM}} (\pi_e; \calD, \hat{q})}
\newcommand{\ips}{\hat{V}_{\mathrm{IPS}} (\pi_e; \calD)}

\newcommand{\dr}{\hat{V}_{\mathrm{DR}} (\pi_e; \calD, \hat{q})}

\newcommand{\mse}{\mathrm{MSE}}
\newcommand{\tpi}{\tilde{\pi}}
\newcommand{\td}{\tilde{\calD}}
\newcommand{\tw}{\tilde{w}}

\definecolor{dkred}{rgb}{0.8,0,0}
\definecolor{dkgreen}{rgb}{0,0.4,0}
\definecolor{tickgreen}{rgb}{0,0.6,0}
\newcommand{\red}[1]{\textcolor{dkred}{#1}}

\begin{document}

\maketitle

\begin{abstract}
Off-policy evaluation (OPE) aims to accurately evaluate the performance of counterfactual policies using only offline logged data. Although many estimators have been developed, there is no single estimator that dominates the others, because the estimators' accuracy can vary greatly depending on a given OPE task such as the evaluation policy, number of actions, and noise level. Thus, the data-driven \textit{estimator selection} problem is becoming increasingly important and can have a significant impact on the accuracy of OPE. However, identifying the most accurate estimator using only the logged data is quite challenging because the ground-truth estimation accuracy of estimators is generally unavailable. This paper thus studies this challenging problem of \emphasize{estimator selection for OPE} for the first time. In particular, we enable an estimator selection that is \textit{adaptive} to a given OPE task, by appropriately subsampling available logged data and constructing \textit{pseudo policies} useful for the underlying estimator selection task. Comprehensive experiments on both synthetic and real-world company data demonstrate that the proposed procedure substantially improves the estimator selection compared to a non-adaptive heuristic.
\end{abstract}

\section{Introduction}
\textit{Off-Policy Evaluation} (OPE) has widely been acknowledged as a crucial technique in search and recommender systems~\citep{gilotte2018offline}. This is because OPE accurately evaluates the performance of counterfactual policies without performing costly A/B tests~\citep{saito2021counterfactual}. This is made possible by leveraging the logged data naturally collected by some \textit{logging} or \textit{behavior} policies. For example, a music recommender system usually records which songs it presented and how the users responded as feedback valuable for estimating the performance of counterfactual policies~\citep{gruson2019offline, kiyohara2022doubly}. Exploiting logged data is, however, often challenging, as the reward is only observed for the chosen action, but not for all the other actions that the system could have taken~\citep{swaminathan2015batch}. Moreover, the logged data is biased due to the distribution shift between the behavior and evaluation policies~\citep{levine2020offline}.

To deal with the difficult statistical estimation involving \textit{counterfactuals} and \textit{distributional shift}, there has been a range of estimators with good theoretical properties -- some estimators ensure unbiasedness under the identification assumptions~\citep{strehl2010learning, precup2000eligibility, dudik2014doubly, jiang2016doubly, thomas2016data}, some reduce the variance while being consistent~\citep{swaminathan2015self, kallus2019intrinsically}, some minimize an upper bound of the mean-squared-error (MSE)~\citep{wang2017optimal, su2020doubly, metelli2021subgaussian}.
Intuitively, having more estimators makes it easier to achieve an accurate OPE. However, this also implies that practitioners now have to carefully solve the \textit{estimator selection} problem to pick the most accurate estimator for their particular task.
If we fail to identify an appropriate estimator, OPE may favor a poor-performing policy that should not be deployed in the field.
Indeed, empirical studies have shown that the estimators' MSE and the most accurate estimator can change greatly depending on task-specific configurations such as the evaluation policy~\citep{voloshin2019empirical}, the size of logged data~\citep{saito2021evaluating}, and the number of actions~\citep{saito2022off}. 
Moreover, \citet{saito2021evaluating} indicate that advanced estimators such as DRos~\citep{su2020doubly} may still produce an inaccurate OPE compared to the typical estimators such as IPS in certain scenarios. These empirical observations suggest the need of an accurate \textit{estimator selection for OPE}.
However, this estimator selection problem has remained completely unaddressed in the existing literature despite its practical relevance.

\begin{figure*}[h]
    \centering
    \begin{minipage}[b]{0.64\linewidth}
      \centering
      \includegraphics[width=1.0\linewidth]{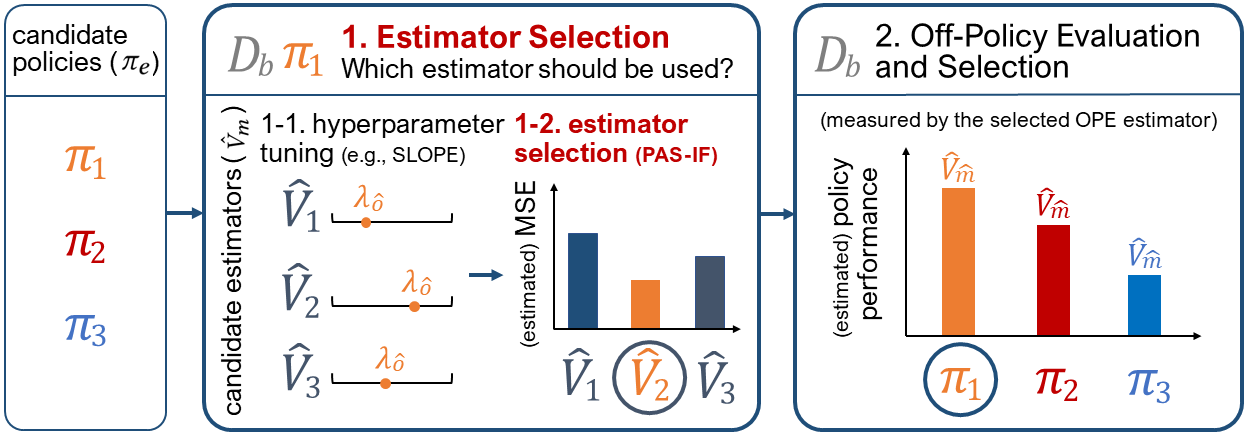}
    \end{minipage}
    \begin{minipage}[b]{0.295\linewidth}
      \centering
      \includegraphics[width=1.0\linewidth]{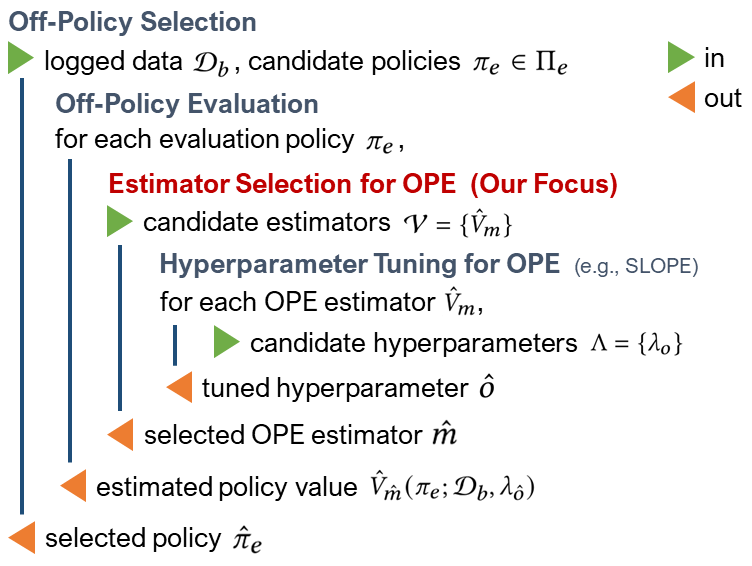}
    \end{minipage}
    \caption{Workflow of Off-Policy Evaluation (OPE) and Selection (OPS) with an Estimator Selection Procesure}
    \label{fig:assumptions}
\end{figure*}

This paper explores the crucial problem of \emphasize{Estimator Selection for OPE} for the first time. 
Specifically, our goal is \emphasize{to select the most accurate estimator among several candidates adaptive to a given OPE task}.
One possible approach to conduct an estimator selection is to first estimate the estimators' MSE using only the logged data and then choose the most accurate. However, estimating the MSE is quite challenging, because it depends on the ground-truth performance of the evaluation policy, which is unavailable to us.
To overcome this issue, we propose a novel estimator selection procedure called \textit{\textbf{P}olicy-\textbf{A}daptive Estimator \textbf{S}election via \textbf{I}mportance \textbf{F}itting} (PAS-IF). 
A key trick lies in PAS-IF is to subsample and divide the logged data into psuedo subpopulations so that one of the subpopulations is deemed to be generated from the evaluation policy.
Once we succeed in creating such subpopulations, we can estimate the candidate estimators' MSE based on the empirical average of the reward in that subpopulation.
PAS-IF synthesizes appropriate subpopulations by minimizing the squared distance between the importance ratio induced by the true evaluation policy and that induced by the pseudo evaluation policy, which we call the \textit{importance fitting} step.
A fascinating feature of our PAS-IF is that it can optimize the subsampling rule differently for different evaluation policies.
In this way, PAS-IF is able to find an accurate estimator \textit{adaptive} to a given OPE task.
This feature is particularly beneficial for the \textit{Off-Policy Policy Selection} (OPS) task where we aim to identify the best policy among several candidates.
Typically, OPS has been solved by applying a single OPE estimator to all candidate policies and picking the best-performing policy based on the OPE results~\citep{doroudi2017importance,kuzborskij2021confident}.
However, our PAS-IF enables us to use the most accurate estimator for each candidate policy, thereby contributing to a much more accurate OPS.

In addition to developing PAS-IF, we empirically compare it to a \textit{non-adaptive} heuristic, which estimates the MSE by naively regarding one of the behavior policies as a pseudo evaluation policy~\citep{saito2020open, saito2021evaluating}. In particular, we demonstrate that our PAS-IF substantially improves the estimator selection accuracy by picking different estimators depending on a given evaluation policy. In contrast, the \textit{non-adaptive} heuristic often fails to identify an accurate estimator when applied to a range of evaluation policies. We also demonstrate that PAS-IF enables a much more accurate OPS compared to the non-adaptive heuristic in both synthetic and real-world experiments.

\section{Related Work}
Here, we summarize some notable existing works.

\paragraph{\textbf{Off-Policy Evaluation}}
OPE has extensively been studied aiming at evaluating counterfactual policies without requiring risky and costly online interactions~\citep{gilotte2018offline, levine2020offline, saito2021counterfactual, kiyohara2021accelerating}. 
Direct Method (DM)~\citep{beygelzimer2009offset}, Inverse Propensity Scoring (IPS)~\citep{strehl2010learning, precup2000eligibility}, and Doubly Robust (DR)~\citep{dudik2014doubly, jiang2016doubly, thomas2016data} are the common baselines for OPE studies. 
DM uses some machine learning algorithms to regress the reward to estimate the policy performance.
DM performs reasonably well when the reward estimation is accurate, however, it is vulnerable to the bias due to model mis-specification~\citep{voloshin2019empirical}. In contrast, IPS enables an unbiased estimation by applying the importance sampling technique. However, IPS can suffer from a high variance particularly when the behavior and evaluation policies deviate greatly~\citep{dudik2014doubly}. DR is a hybrid of DM and IPS, and often reduces the variance of IPS while remaining unbiased. However, DR can still suffer from a high variance especially when the action space is large~\citep{saito2022off}. With the goal of achieving a better bias-variance trade-off, researchers have produced a number of estimators~\citep{wang2017optimal, kallus2019intrinsically, su2020doubly}. 
For example, \citet{su2019cab} combine DM and IPS via adaptive weighting while \citet{metelli2021subgaussian} modify the importance ratio to improve the typical exponential concentration rate of the error bound to a subgaussian rate. 
We build on the vast literature on OPE, but rather focus on the relatively under-explored problem of selecting the most accurate estimator among many options in a data-driven way.

\paragraph{\textbf{Hyperparameter Tuning for OPE}}
Hyperparameter tuning (which is fundamentally different from our estimator selection task) often plays a crucial role in OPE. The bias-variance trade-off of many OPE estimators such as Switch~\citep{wang2017optimal} and DRos~\citep{su2020doubly} depends heavily on their built-in hyperparameters~\citep{saito2020open,saito2021evaluating}. A naive way to tune such hyperparameters is to estimate the bias and variance of a given estimator and construct some MSE surrogates~\citep{thomas2016data, wang2017optimal, su2020doubly}. However, estimating the bias is equally difficult as OPE itself because the bias depends on the ground-truth performance of the evaluation policy~\citep{su2020adaptive}. 
To tackle this issue, \citet{su2020adaptive} and \citet{tucker2021improved} propose a procedure called SLOPE based on the Lepski's principle, which was originally proposed for non-parametric bandwidth selection~\citep{lepski1997optimal}. Since SLOPE avoids estimating the bias, it theoretically and empirically works better than the naive tuning procedure relying on some MSE surrogates~\citep{su2020adaptive,tucker2021improved}.
However, SLOPE is only applicable to the hyperparameter tuning of a particular estimator class due to its \textit{monotonicity} assumption.
In contrast, our estimator selection problem aims at comparing different estimator classes and select the most accurate one, possibly after applying SLOPE within each estimator class.\footnote{For example, SLOPE can be used to tune the hyperparameters of Switch and DRos, however, it does not tell us which estimator is better. Our interest is thus to select the better estimator class.}

\paragraph{\textbf{Estimator Selection for OPE}} 
Given several candidate OPE estimators, the estimator selection problem aims to identify the most accurate one. Although this problem has remained unexplored in the existing literature, many empirical studies have shown that the quality of estimator selection can have a significant impact on the accuracy of the downstream OPE. 
For instance, \citet{voloshin2019empirical} thoroughly investigate the estimation accuracy of OPE estimators in a range of simulated environments. Their empirical results demonstrate that the estimators' MSE and the most accurate estimator can change greatly depending on the data size, divergence between the behavior and evaluation policies, and many other environmental configurations.
\citet{saito2021evaluating} also demonstrate that recent estimators with better theoretical guarantees can sometimes produce an inaccurate OPE compared to the typical estimators, suggesting the necessity of identifying and using appropriate estimators adaptive to a given OPE task.
We are the first to formally formulate the important problem of \textit{estimator selection for OPE}.
We also propose a method to enable an \textit{adaptive} estimator selection and empirically demonstrate that the accuracy of estimator selection makes a non negligible difference in the accuracy of the downstream OPE and OPS.

\paragraph{\textbf{Off-Policy Policy Selection}}
OPS aims to select the best-performing \textit{policy} among several candidate policies using only the logged bandit data~\citep{paine2020hyperparameter, tang2021model, zhang2021towards}.
When applied to OPS, OPE estimators should work reasonably well among a range of candidate evaluation policies.
However, this desideratum is often hard to achieve, because an estimator's accuracy usually vary substantially when applied to different evaluation policies~\citep{voloshin2019empirical}. 
Existing works on OPS deal with this instability of OPE by leveraging some high probability bounds on policy value estimates~\citep{thomas2015confidence, thomas2015high, kuzborskij2021confident, hao2021bootstrapping, yang2020offline}. In particular, \citet{doroudi2017importance} validate whether a fair policy comparison is possible based on a concentration inequality. \citet{yang2021pessimistic} estimate a pessimistic policy performance to alleviate an overestimation, which could lead to an inaccurate OPS.
However, there is no existing work attempting to \textit{switch} estimators adaptive to each evaluation policy among the set of candidates.
As a potential application of our proposed \textit{estimator selection} procedure, we show, in our experiments, that adaptive estimator selection can also significantly improve the accuracy of the \textit{policy selection} task.

\section{Problem Formulation}\label{sec:setup}

This section first formulates OPE of contextual bandits and then the corresponding estimator selection problem, which is our primary interest.

\subsection{Off-Policy Evaluation (OPE)}

Let $x \in \calX$ be a context vector (e.g., user demographics) that the decision maker observes when choosing an action. 
Let $r \in [0, r_{max}]$ be a reward (e.g., whether a coupon assignment results in an increase in revenue). Context and reward are sampled from some unknown distributions $p(x)$ and $p(r | x, a)$, where $a \in \calA$ is a discrete action (i.e., a coupon). We call a function $\pi: \calX \rightarrow \Delta(\calA)$ a policy, where $\pi(a | x)$ is the probability of taking action $a$ given context $x$.

In OPE, we are interested in accurately estimating the following \textit{(policy) value}:
\begin{align*}
    V(\pi) := \mE_{p(x)\pi(a|x)p(r|x,a)}[r].
\end{align*}
The most reliable way to estimate the policy value of \textit{evaluation policy} $\pi_e$ is to actually deploy $\pi_e$ in an online environment (a.k.a. A/B tests). 
However, such an \textit{on-policy} evaluation is often limited in practice, as it incurs large implementation costs~\citep{matsushima2021deployment} and there is the risk of deploying poor policies~\citep{gilotte2018offline}. Therefore, it is often desirable to evaluate the policy value via OPE at first and then pick only a small number of promising policies to be evaluated via A/B tests~\citep{irpan2019off}.

To estimate the policy value, OPE leverages the logged bandit data $\calD_b := \{(x_i, a_i,r_i)\}_{i=1}^n$ collected by a behavior policy as follows.
\begin{align*}
  \{(x_i,a_i,r_i)\}_{i=1}^n \ & \sim \ \prod_{i=1}^n p(x_i) \underbrace{p(j_i)\pi_{j_i}(a_i|x_i)}_{\pi_b (a_i | x_i)} p(r_i | x_i, a_i),
\end{align*}
where $\pi_b$ is the behavior policy, which may consist of $l \,(\geq 1)$ different data collection policies $\pi_1, \cdots, \pi_l$.\footnote{This is a general formulation, including the standard setting with a single data collection policy ($l=1$) as a special case. Moreover, our setting is fundamentally different from the multiple logger setting of~\citet{agarwal2017effective,kallus2020optimal}, which assume the deterministic behavior policy assignment.}
Here, $\calD_b = \bigcup_{j=1}^l \calD_j$ can be seen as an aggregate of several logged datasets, each of which contains $n_j$ observations collected by the $j$-th data collecting policy $\pi_j$.

The goal of OPE is then to estimate the aforementioned policy value of evaluation policy using only the logged data: $V(\pi_e) \approx \hat{V}(\pi_e; \calD_b)$. The accuracy of an estimator $\hat{V}$ is typically quantified by the \textit{mean-squared-error} (MSE):
\begin{align}
    & \mse(\hat{V}; \pi_e, \pi_b, n) \\
    &:= \mE_{\calD_b} \left[ (\hat{V}(\pi_e; \calD_b) - V(\pi_e) )^2 \right] \nonumber \\
    &= (\mathrm{Bias}(\hat{V}; \pi_e, \pi_b, n))^2 + \mV_{\calD_b} (\hat{V}; \pi_e, \pi_b, n), \label{eq:mse}
\end{align}
As suggested in Eq.~\eqref{eq:mse}, achieving a reasonable bias-variance tradeoff is critical in enabling an accurate OPE. 
This motivates many estimators to be developed, including DM~\citep{beygelzimer2009offset}, IPS~\citep{precup2000eligibility}, DR~\citep{dudik2014doubly}, Switch~\citep{wang2017optimal}, DRos~\citep{su2020doubly}, and DR-$\lambda$~\citep{metelli2021subgaussian}.\footnote{We provide the definition and important statistical properties of these estimators in Appendix~\ref{app:estimators}.}  
However, as we have already argued, many empirical studies imply that there is no estimator that is universally the best~\citep{voloshin2019empirical,saito2020open,saito2021evaluating}. This empirical evidence leads us to study the data-driven \textit{estimator selection} problem for OPE, which we describe in detail below.

\subsection{Estimator Selection for OPE} \label{sec:estimator_selection}
The goal of estimator selection is to select the most accurate estimator (which may change depending on a given OPE task) from a candidate pool of estimator classes $\mathcal{V}:=\{ \hat{V}_m \}_{m=1}^M$.
An ideal strategy for this estimator selection task would be to pick the estimator achieving the lowest MSE:
\begin{align}
    m^{\ast} := \argmin_{m \in \{1,\ldots,M\}} \,\mse(\hat{V}_m; \pi_e, \pi_b, n). \label{eq:estimator_selection}
\end{align}
Unfortunately, however, Eq.~\eqref{eq:estimator_selection} is infeasible because the MSE depends on the policy value of the evaluation policy, which is arguably unknown.
Therefore, we instead consider performing an \textit{offline} estimator selection based on an estimated MSE.
\begin{align*}
    \hat{m} := \argmin_{m \in \{1,\ldots,M\}} \,\widehat{\mse}(\hat{V}_m; \calD_b). 
    % \label{eq:off_policy_estimator_selection}
\end{align*}
Although there is no existing literature that formally discusses this estimator selection problem, the following describes a (non-adaptive) heuristic as a reasonable baseline.

\paragraph{\textbf{Non-Adaptive Heuristic}}
One possible approach to estimate the MSE is to naively regard one of the data collection policies $\pi_j$ as a \textit{pseudo} evaluation policy. Then, the \textit{non-adaptive} heuristic estimates the MSE of a given estimator $\hat{V}$ as follows.\footnote{This heuristic has been used in some empirical studies of OPE~\citep{saito2020open, saito2021data, saito2021evaluating}, however, the estimator selection problem itself has not yet been formally formulated in the OPE literature.}
\begin{align*}
    & \widehat{\mse}(\hat{V}; \calD_b) \\
    &:= \frac{1}{|\calS|} \sum_{s \in \calS} \left(\hat{V}(\pi_j^{(s)}; \calD_{b \setminus j}^{\ast (s)}) - V_{\mathrm{on}}(\pi_j^{(s)}; \calD_j^{(s)}) \right)^2, % \label{eq:baseline}
\end{align*}
where $\pi_j$ is a pseudo evaluation policy.  $V_{\mathrm{on}}(\pi_j; \calD_j) := \sum_{i=1}^{n_j} r_i / n_j$ is its on-policy policy value estimate. $\calD_{b \setminus j} := \calD_b \setminus \calD_j$ is the logged data collected by the corresponding (pseudo) behavior policy. $\calD^{\ast}_b$ indicates bootstrapped samples of $\calD_b$, and $\calS$ is a set of random seeds for bootstrap. $\pi_j$ is either randomly picked among available data collection policies~\citep{saito2021evaluating} or is fixed~\citep{saito2020open, saito2021data} for every random seed.

A critical pitfall of this heuristic is that it cannot accurately estimate the MSE when the data collection policies (one of which is used as the pseudo evaluation policy) are totally different from $\pi_e$.
In fact, as we will show in the synthetic experiment, the non-adaptive heuristic often fails to choose an appropriate OPE estimator when there is a large divergence between the true evaluation policy and the pseudo evaluation policy.
Unfortunately, such an undesirable situation is often the case in practice, because we usually want to evaluate counterfactual policies that have never been deployed. This motivates us to develop a novel \textit{adaptive} estimator selection procedure that can estimate the estimators' MSE by taking the task-specific configurations (such as the evaluation policy) into account.

\section{Our Adaptive Approach} \label{sec:pasif}
This section proposes a new estimator selection procedure called
\textit{\textbf{P}olicy-\textbf{A}daptive Estimator \textbf{S}election via \textbf{I}mportance \textbf{F}itting} (PAS-IF).

\begin{figure}
    \centering
    \includegraphics[width=0.95\linewidth]{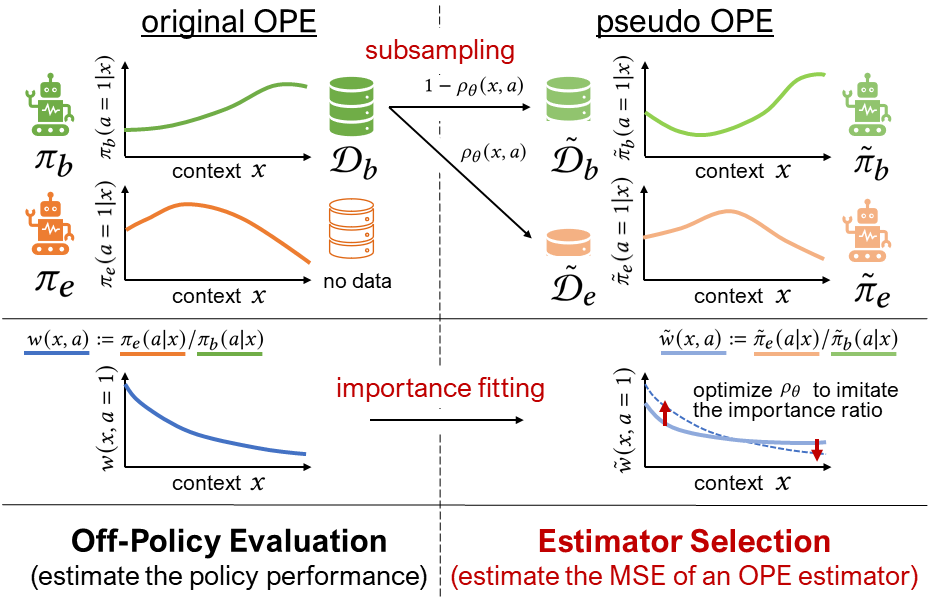}
    \caption{High level overview of \textit{Policy-Adaptive Estimator Selection via Importance Fitting} (PAS-IF)}
    \label{fig:pasif}
\end{figure}

\begin{figure}[tb]
\centering
\begin{algorithm}[H]
\caption{\textbf{P}olicy-\textbf{A}daptive Estimator \textbf{S}election via \textbf{I}mportance \textbf{F}itting (PAS-IF)}
\begin{algorithmic}[1]
\REQUIRE a candidate set of OPE estimators $\mathcal{V}$, logged bandit dataset $\calD_b$, evaluation policy $\pi_e$, target partition rate $k$, learning rate $\eta$, regularization coefficient $\lambda$, maximum steps $T$, a set of random seeds $\calS$
\ENSURE selected OPE estimator $\hat{V}_{\hat{m}}$
\FOR{$m = 1,\ldots,M$}
\FOR{$s \in \calS$} 
    \STATE $ \calD_b^{\ast(s)} \leftarrow \mathrm{Bootstrap} (\calD_b; s) $
    \STATE Initialize subsampling parameters $\theta$
    \FOR{$t = 1, \ldots, T$}
        \STATE $ \theta \leftarrow \theta - \eta \left( \frac{\partial D}{ \partial \theta} + \lambda \frac{\partial R}{ \partial \theta} \right) \quad \triangleright \mathrm{importance \, fitting}$
    \ENDFOR
    \STATE $ \td_e^{\ast(s)}, \td_b^{\ast(s)} \leftarrow \mathrm{Subsample}(\calD_b^{\ast(s)}; \rho_{\theta}) $
    \STATE $z_s \leftarrow (\hat{V}_m(\tpi_e; \td_b^{\ast(s)}) - V_{\mathrm{on}}(\tpi_e; \td_e^{\ast(s)}))^2 $
\ENDFOR
\STATE $ \widehat{\mse}(\hat{V}_m; \calD_b) \leftarrow \sum_{s \in \calS} z_s / |\calS| $
\ENDFOR
\STATE $ \hat{m} \leftarrow \argmin_{m \in \{1, \ldots, M\}} \widehat{\mse}(\hat{V}_m; \calD_b) $
\end{algorithmic}
\label{algo:PAS-IF}
\end{algorithm}
\end{figure}

Our key idea is to subsample the logged data and generate subpopulations to accurately estimate the MSE for a range of evaluation policies. 
For this, we introduce a subsampling rule $\rho_{\theta}: \calX \times \calA \rightarrow (0, 1)$ parameterized by $\theta$. $\rho_{\theta}$ allocates each observation in $\calD_b$ to pseudo evaluation dataset $\td_e$ or pseudo behavior dataset $\td_b$ where $ \rho_{\theta}(x,a) \in (0,1)$ is the probability of the data $(x,a,\cdot)$ being allocated to $\td_e$. Under this formulation, the pseudo datasets can be seen as generated by the following pseudo policies:
\begin{align*}
    \tpi_e(a|x) &:= \pi_b(a|x) \frac{\rho_{\theta}(x,a)}{\mE_{\pi_b(a|x)}[\rho_{\theta}(x,a)]}, \\
    \tpi_b(a|x) &:= \pi_b(a|x) \frac{1 - \rho_{\theta}(x,a)}{1 - \mE_{\pi_b(a|x)}[\rho_{\theta}(x,a)]},
\end{align*}

This formulation allows us to control the data generation process of $\td_e$ and $\td_b$ \textit{adaptive} to a given OPE task by appropriately optimizing the subsampling rule $\rho_{\theta}$. In particular, when optimizing $\rho_{\theta}$, we try to imitate the true importance ratio $w(x,a):=\pi_e(a|x)/\pi_b(a|x)$, as it plays a significant role in determining the bias-variance tradeoff of OPE~\citep{voloshin2019empirical}.
Specifically, we minimize the following squared distance between the importance ratio induced by the true policies and that induced by the pseudo policies:
\begin{align*}
    D(\pi,\tpi) &:= \mE_{p(x)\pi_b(a|x)} [(w(x,a)-\tw(x,a))^2 ],
\end{align*}
where we define $\tw(x,a):=\tpi_e(a|x)/\tpi_b(a|x)$. We minimize this \textit{importance fitting} objective by gradient decent where the gradient of $d(x, a):=w(x,a)-\tw(x,a)$ is given as:\footnote{Appendix~\ref{app:deviation} provides how we derive these gradients.}
\begin{align*}
    \frac{\partial d(x,a)}{\partial \rho_{\theta}}
    & \propto \frac{(\tw(x,a) - w(x,a))}{(1-\rho_{\theta}(x,a))^2} \left( \frac{1}{\mE_{\pi_b}[\rho_{\theta}(x,a)]} - 1 \right) \\
    & \quad \; \cdot \left( 1 - \frac{\pi_b(x,a) \rho_{\theta}(x,a) (1 - \rho_{\theta}(x,a))}{(\mE_{\pi_b}[\rho_{\theta}(x,a)])(1 - \mE_{\pi_b}[\rho_{\theta}(x,a)])} \right),
\end{align*}

This importance fitting step enables PAS-IF to estimate the MSE adaptive to a range of evaluation policies while preserving a bias-variance tradeoff of a given OPE task, thereby improving the quality of estimator selection compared to the non-adaptive heuristic.
A potential concern is that the size of the pseudo behavior dataset $\td_b$ may deviate greatly from that of the original logged data $\calD_b$ depending on $\rho_{\theta}$. 
If the size of $\td_b$ is much smaller than that of $\calD_b$, PAS-IF may prioritize the variance more than necessary when performing estimator selection. 
To alleviate this potential failure, we set a target partition rate $k$ and apply the following regularization when optimizing $\rho_{\theta}$.
\begin{align*}
    R(\tpi, k) := \mE_{p(x)} \left[ \left( \mE_{\pi_b(a|x)}\left[ \rho_{\theta}(x,a) \right] - k \right)^2 \right],
\end{align*}
where we impose regularization on every context to preserve $p(x)$ between $\td_e$ and $\td_b$.

Combining the importance fitting objective $D(\cdot)$ and regularization $R(\cdot)$, our PAS-IF optimizes the subsampling rule $\rho_{\theta}$ by iterating the following gradient update.
\begin{align*}
    \theta 
    & \leftarrow \theta - \eta \left( \frac{\partial D}{ \partial \theta} + \lambda \frac{\partial R}{ \partial \theta} \right),
\end{align*}
where $\lambda$ is a regularization coefficient and $\eta$ is a learning rate. 
Note that, in our experiments, we pick the regularization coefficient $\lambda$ so that $\mE_{\pi_b(a|x)}\left[ \rho_{\theta}(x,a) \right] \in [0.18,0.22]$. By doing this, we ensure that $|\td_b|$ becomes sufficiently close to $|\calD_b|$, while making $V_{\mathrm{on}}(\pi_e;\td_e) \approx V(\pi_e)$ reasonably accurate. We later demonstrate that this heuristic tuning procedure works reasonably well in a range of settings.

Once we optimize $\rho_{\theta}$ for a given OPE task, we subsample the logged data to create pseudo datasets and estimate the MSE of an estimator $\hat{V}$ as follows.
\begin{align*}
    &\widehat{\mse}(\hat{V}; \calD_b) \\
    &:= \frac{1}{|\calS|} \sum_{s \in \calS} \left(\hat{V}(\tpi_e^{(s)}; \td_b^{\ast (s)}) - V_{\mathrm{on}}(\tpi_e^{(s)}; \td_e^{\ast (s)}) \right)^2.
\end{align*}
where $\calS$ is a set of random seeds for bootstrap sampling, and $ \calD_b^{\ast(s)}$ is the $s$-th bootstrapped logged dataset sampled from $\calD_b$. $\td_e^{\ast (s)}$ and $\td_b^{\ast (s)}$ are subsampled from $ \calD_b^{\ast(s)}$ based on $\tpi_e^{(s)}$ and $\tpi_b^{(s)}$, respectively. This bootstrapping procedure aims to stabilize PAS-IF.
Algorithm~\ref{algo:PAS-IF} summarizes the whole estimator selection procedure based on PAS-IF.\footnote{In addition to the primary benefit (i.e., \textit{adaptive} estimator selection), PAS-IF is also able to relax some assumptions about the data collection policies compared to the non-adaptive heuristic. Appendix~\ref{app:additional_benefit} discusses this additional benefit of PAS-IF in detail.}

\section{Synthetic Experiments}
This section compares our PAS-IF with the non-adaptive heuristic in terms of estimator selection and OPS. 
Note that our synthetic experiment is implemented on top of \textit{OpenBanditPipeline}~\citep{saito2020open}.\footnote{https://github.com/st-tech/zr-obp} Our experiment code is available at \textbf{https://github.com/sony/ds-research-code/tree/master/aaai23-pasif} and Appendix~\ref{app:experiment_detail} describes some additional experiment details. 

\subsection{Setup} \label{sec:synthetic_setup}

\paragraph{Basic setting.} To generate synthetic data, we first randomly sample 10-dimensional context $x$, independently and normally distributed with zero mean. We also set $|\calA|=10$. The binary rewards $r$ are sampled from the Bernoulli distribution as $r \sim Bern(q(x,a))$ where $q(x,a)$ is \texttt{obp.dataset.logistic\_reward\_function}.

\paragraph{\textbf{Data Collection}} 
We define our behavior policy $\pi_b$ based on the two different data collection policies $\pi_1$ and $\pi_2$:
\begin{align}
    \pi_j(a|x) := \frac{\exp(\beta_j \cdot q(x, a))}{\sum_{a' \in \calA} \exp(\beta_j \cdot q(x, a'))}, \label{eq:policy}
\end{align}
where $j \in \{1,2\}$ and $q(x, a) := \mE[r|x,a]$ is the expected reward.
$\beta_j$ is an inverse temperature parameter of the softmax function. A positive value of $\beta$ leads to a near-optimal policy, while a negative value leads to a bad policy. When $\beta = 0$, $\pi_j$ is identical to uniform random. 
The logged dataset contains $n = 2,000$ observations with $p(j=1) = p(j=2) = 1/2$, and we try two different sets of data collection policies: (i) $(\beta_1, \beta_2) = (-2, 2)$ and (ii) $(\beta_1, \beta_2) = (3, 7)$.

\paragraph{\textbf{Evaluation Policies}} 
We also follow Eq.~\eqref{eq:policy} to define evaluation policies and vary $\beta_e \in \{ -10, -9, \cdots, 10 \}$ in the estimator selection task to evaluate the estimator selection accuracy for various evaluation policies. For the policy selection task, we prepare 20 candidate policies learned by different policy learning methods to make OPS reasonably hard. We describe how to define the candidate policies for the policy selection experiment in detail in Appendix~\ref{app:experiment_detail}.

\paragraph{\textbf{Candidate OPE Estimators}} 
Table~\ref{tab:estimators} summarizes the candidate set of estimator classes ($\mathcal{V}$) containing $M=21$ candidate OPE estimators with varying reward predictors $\hat{q}$. Note that we perform SLOPE~\citep{su2020adaptive,tucker2021improved} to tune each estimator's built-in hyperparameter before performing estimator selection to simulate a realistic situation where we combine SLOPE and PAS-IF to improve the end-to-end OPE pipeline.

\begin{table}[tb]
\large
\centering
\caption{Candidate OPE estimators} \label{tab:estimators}
\vspace{-2mm}
\scalebox{0.7}{
\begin{tabular}{c|c}
    \toprule
    OPE estimator & Choices of $\hat{q}$ \\ \midrule \midrule
    DM & $\{ \rm{RandomForest}, \rm{LightGBM} , \rm{LogisticRegression} \}$ \\
    IPSps & - \\
    DRps & $\{ \rm{RandomForest}, \rm{LightGBM} , \rm{LogisticRegression} \}$ \\
    SNIPS & - \\
    SNDR & $\{ \rm{RandomForest}, \rm{LightGBM} , \rm{LogisticRegression} \}$ \\
    Switch & $\{ \rm{RandomForest}, \rm{LightGBM} , \rm{LogisticRegression} \}$ \\
    DRos & $\{ \rm{RandomForest}, \rm{LightGBM} , \rm{LogisticRegression} \}$ \\
    IPS-$\lambda$ & - \\
    DR-$\lambda$ & $\{ \rm{RandomForest}, \rm{LightGBM} , \rm{LogisticRegression} \}$ \\
\bottomrule
\end{tabular}
}
\vskip 0.1in
\raggedright
\fontsize{8.0pt}{8.0pt}\selectfont \textit{Note}: We use 3-fold cross-fitting~\citep{narita2021debiased} when training $\hat{q}$. We also use SLOPE~\citep{su2020adaptive,tucker2021improved} to tune the estimators' built-in hyperparemeters. Appendix~\ref{app:estimators} defines and describes these OPE estimators in detail.
\end{table}

\paragraph{\textbf{Compared Methods}}
We compare PAS-IF with the non-adaptive heuristic.  
For PAS-IF, we set $\calS= \{ 0, 1, \ldots, 9 \}$, $k=0.2$, $\eta = 0.001$, and $T = 5,000$, and select the regularization coefficient $\lambda$ from $\{ 10^{-1}, 10^0, 10^1, 10^2, 10^3 \}$ by a procedure described in Section~\ref{sec:pasif}.

\paragraph{\textbf{Evaluation Metrics}} 
We quantify the estimator selection accuracy of PAS-IF and the non-adaptive heuristic by the following metrics.
\begin{itemize}
    \item \textbf{Relative Regret (e)} (lower is better): This evaluates the accuracy of the estimator selected by each method compared to the most accurate estimator.
    \begin{align*}
        &\mathrm{rRegret}^{(e)} \\
        &:= \frac{\mse(\hat{V}_{\hat{m}}; \pi_e, \pi_b, n)) - \mse(\hat{V}_{m^{\ast}}; \pi_e, \pi_b, n)}{ \mse(\hat{V}_{m^{\ast}}; \pi_e, \pi_b, n)}.
    \end{align*}
    \item \textbf{Rank Correlation (e)} (higher is better): This is the Spearman's rank correlation between the true and estimated MSE. This metric evaluates how well each method preserves the ranking of the candidate estimators in terms of their MSE.
\end{itemize}

In addition to the above metrics regarding estimator selection, we also evaluate PAS-IF and the non-adaptive heuristic in terms of the quality of the resulting OPS. Note that OPS is an important application of estimator selection whose aim is to select the best-performing policy based on OPE as:
\begin{align*}
    \hat{\pi}_e := \argmax_{\pi_e \in \Pi_e} \hat{V}_{\hat{m}}(\pi_e; \calD_b),
\end{align*}
where $\hat{V}_{\hat{m}}$ is the OPE estimator selected either by PAS-IF or the non-adaptive heuristic.
We can evaluate the estimator selection methods with respect to their resulting OPS quality using the following metrics~\citep{paine2020hyperparameter}.
\begin{itemize}
    \item \textbf{Relative Regret (p)} (lower is better): This metric measures the performance of the \textit{policy} $\hat{\pi}_e$ selected based on PAS-IF or the non-adaptive heuristic compared to the best policy $\pi_e^{\ast}$ among the candidates in $\Pi_e$. $$\mathrm{rRegret}^{(p)} := \frac{V(\pi_e^{\ast}) - V(\hat{\pi}_e)}{V(\pi_e^{\ast})}.$$
    \item \textbf{Rank Correlation (p)} (higher is better): This is the Spearman's rank correlation between the ground-truth performance of the candidate policies ($\{V(\pi_e)\}_{\pi_e \in \Pi_e}$) and those estimated by the selected estimator ($\{\hat{V}_{\hat{m}}(\pi_e)\}_{\pi_e \in \Pi_e}$). 
\end{itemize}

\begin{figure*}[htbp]
  \centering
  \begin{minipage}[b]{0.40\linewidth}
  \centering
  \includegraphics[width=0.9\linewidth]{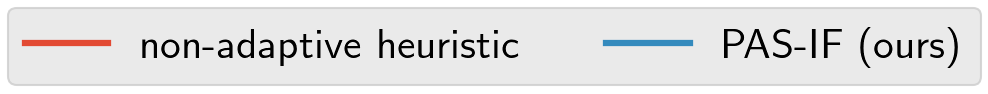}
  \vspace{2mm}
  \end{minipage} \\
  \begin{minipage}[b]{0.85\linewidth}
   \centering
      \begin{minipage}[b]{0.45\linewidth}
        \includegraphics[width=1.05\linewidth]{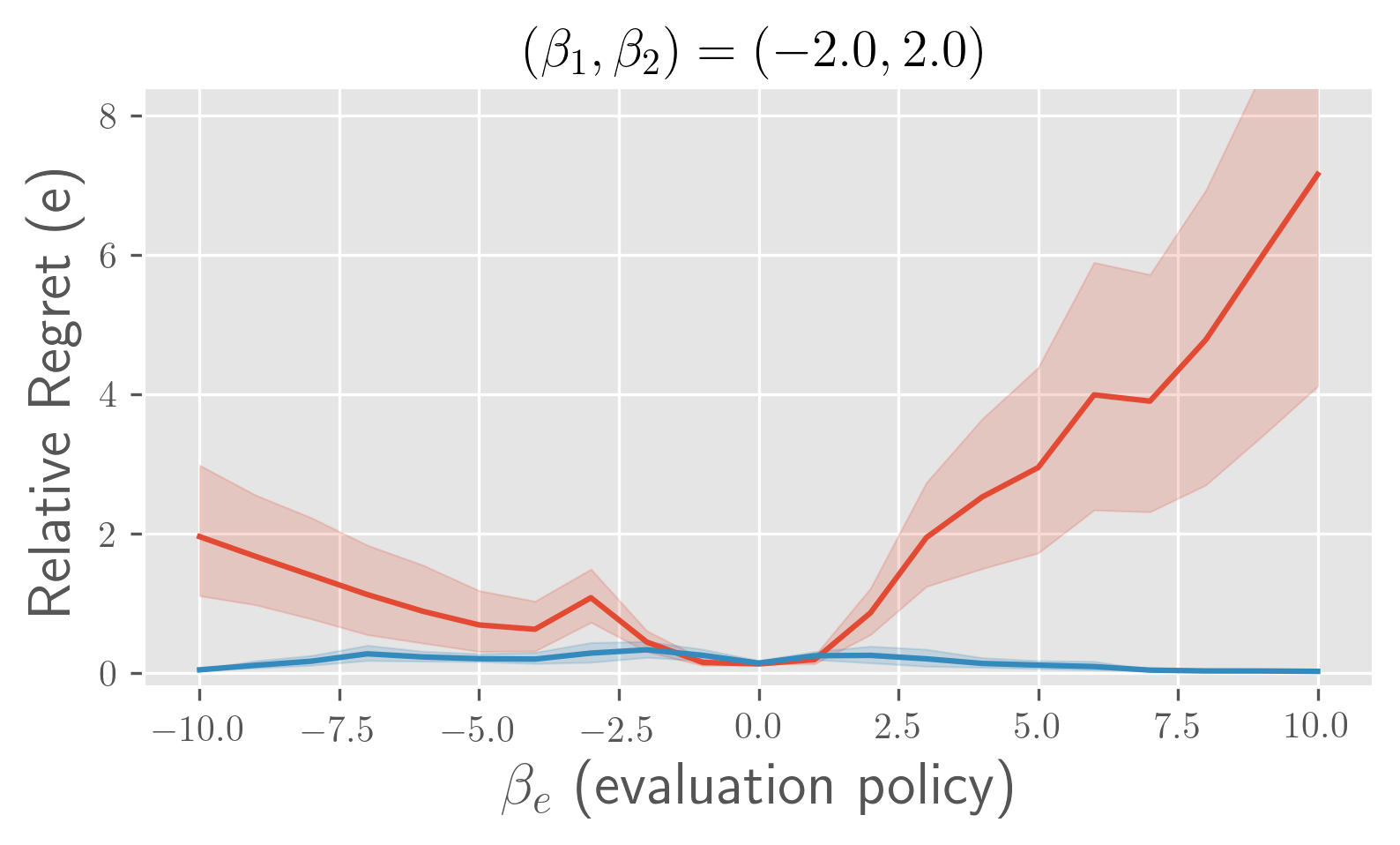}
      \end{minipage}
      \begin{minipage}[b]{0.45\linewidth}
        \includegraphics[width=1.05\linewidth]{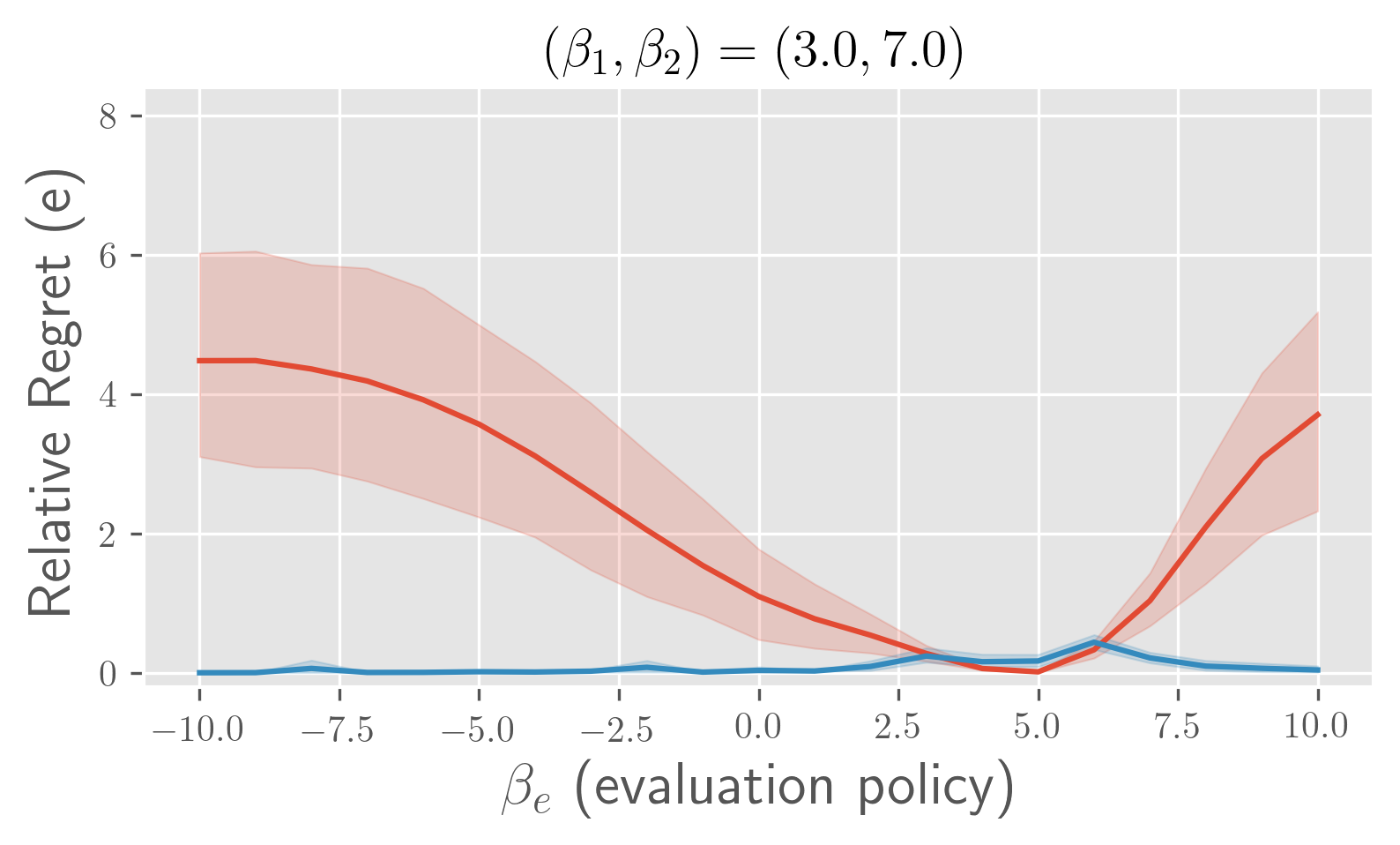}
      \end{minipage}
      \vspace{-1mm}
      \caption{\textbf{Relative Regret (e)} with varying evaluation policies ($\beta_e$) in the synthetic experiment.}
      \vspace{1mm}
      \label{fig:regret}
  \end{minipage}
\end{figure*}

\begin{table*}[htb]
\large
\centering
\vspace{2mm}
\caption{\textbf{Relative Regret (e)} and \textbf{Rank Correlation (e)} in estimator selection in the real-world experiment.} \label{tab:real}
\def\arraystretch{1.2}
\scalebox{0.68}{
\begin{tabular}{c|cc|cc|cc}
\toprule
evaluation policy & \multicolumn{2}{c}{$j=1$} & \multicolumn{2}{c}{$j=2$} & \multicolumn{2}{c}{$j=3$} \\ \midrule
metric & \textbf{Relative Regret (e)} & 
\textbf{Rank Correlation (e)} & \textbf{Relative Regret (e)} & 
\textbf{Rank Correlation (e)} & \textbf{Relative Regret (e)} & 
\textbf{Rank Correlation (e)} \\\midrule \midrule
non-adaptive heuristic
& 1.755 \, ($\pm$ 1.032)
& -0.034 \, ($\pm$ 0.507)
& 1.827 \, ($\pm$ 3.045)
& 0.192 \, ($\pm$ 0.374)
& 3.580 \, ($\pm$ 1.676)
& 0.015 \, ($\pm$ 0.231)
\\
PAS-IF (ours)
& \textbf{1.609} \, ($\pm$ 0.668)
& \textbf{0.002} \, ($\pm$ 0.286)
& \textbf{0.691} \, ($\pm$ 0.708)
& \textbf{0.342} \, ($\pm$ 0.546)
& \textbf{1.889} \, ($\pm$ 1.325)
& \textbf{0.124} \, ($\pm$ 0.312)
\\
\bottomrule
\end{tabular}
}
\vskip 0.1in
\raggedright
\fontsize{9pt}{9pt}\selectfont \textit{Note}:
A lower value is better for Relative Regret (e), while a higher value is better for Rank Correlation (e). 
The \textbf{bold} fonts represent the best estimator selection method. The values in parentheses indicate the standard deviation of the metrics estimated over 10 experiment runs.
\end{table*}

\subsection{Result} Figure~\ref{fig:regret} shows Relative Regret (e) in estimator selection where the results are averaged over 100 simulations performed with different seeds.
The result clearly demonstrates that our PAS-IF substantially improves relative regret compared to the non-adaptive heuristic for a range of evaluation policies ($\beta_e$).
In particular, we observe that PAS-IF is able to identify an accurate OPE estimator (that has a low regret) even in cases where the behavior and evaluation policies deviate greatly (i.e., $\beta_e$ is different from $\beta_1$ and $\beta_2$), while the non-adaptive heuristic fails dramatically. 
We also observe the similar trends in terms of the rank correlation metric, which we report in Appendix~\ref{app:experiment_detail}. These results demonstrate that being adaptive to a given OPE task such as evaluation policy is crucial for effective estimator selection.

Table~\ref{tab:ops} compares PAS-IF and the non-adaptive heuristic in terms of their OPS quality when $(\beta_1,\beta_2)=(-2.0,2.0)$.\footnote{We observe the similar results when $(\beta_1,\beta_2)=(3.0,7.0)$, which is reported in Table~\ref{tab:ops3-7} in the appendix.} The results indicate that PAS-IF is better in terms of both relative regret and rank correlation. Moreover, PAS-IF has a smaller standard deviation for both metrics, suggesting its stability. These observations indicate that an adaptive estimator selection also has a substantial positive benefit on the downstream OPS task.

\begin{table}[htb]
\large
\centering
\caption{\textbf{Relative Regret (p)} and \textbf{Rank Correlation (p)} in OPS in the synthetic experiment: $(\beta_1,\beta_2)=(-2.0,2.0)$.} \label{tab:ops}
\def\arraystretch{1.2}
\scalebox{0.725}{
\begin{tabular}{c|cc}
\toprule
 & \textbf{Relative Regret (p)} & 
\textbf{Rank Correlation (p)} \\\midrule \midrule
non-adaptive heuristic
& 0.0341 \, ($\pm$ 0.0626)
& 0.7452 \, ($\pm$ 0.6211)
\\
PAS-IF (ours)
& \textbf{0.0077} \, ($\pm$ 0.0144)
& \textbf{0.9619} \, ($\pm$ 0.0210)
\\
\bottomrule
\end{tabular}
}
\vskip 0.1in
\raggedright
\fontsize{9pt}{9pt}\selectfont \textit{Note}:
A lower value is better for Relative Regret (p), while a higher value is better for Rank Correlation (p).
\end{table}

\section{Real-World Experiment}
In this section, we apply PAS-IF to a real e-commerce application and demonstrate its practical benefits.

\paragraph{\textbf{Setup}}
For this experiment, we conducted a data collection A/B test in April 2021 on an e-commerce platform whose aim is to optimize a coupon assignment policy that facilitates user consumption. Thus, $a$ is a coupon assignment variable where there are six different types of coupons ($|\calA|=6$). $r$ is the revenue within the 7-day period after the coupon assignment. During data collection, we deployed three different coupon assignment policies ($\pi_1$, $\pi_2$, and $\pi_3$) and assign them (almost) equally to the users, resulting in $n=40,985$.\footnote{$p(j=1) \approx p(j=2) \approx p(j=3) \approx 1/3$.}

We compare PAS-IF with the non-adaptive heuristic for three different settings ($j'=1,2,3$). Each setting regards $\pi_{j'}$ as the evaluation policy and $\calD_b = \calD \setminus \calD_{j'}$ as the logged data. For example, when we consider $\pi_1$ as an evaluation policy, $\calD_b$ is $\calD_2 \cup \calD_3$. Then, we use $V_{\mathrm{on}}(\pi_{j'}; \calD_{j'})$ (which is not available to PAS-IF and the non-adaptive heuristic) to estimate the true MSE of an estimator $\hat{V}$ as follows.
\begin{align*}
    \mse_{\mathrm{on}}(\hat{V}; \calD)
    &:= \frac{1}{|\calS|} \sum_{s \in \calS} \left(\hat{V}(\pi_{j'}; \calD_b^{\ast (s)}) - V_{\mathrm{on}}(\pi_{j'}; \calD_{j'}) \right)^2,
\end{align*}
where $\calS := \{ 0, \ldots, 9 \}$ is a set of random seeds for bootstrap sampling. We evaluate the estimator selection performance of PAS-IF and the non-adaptive heuristic by calculating the metrics regarding estimator selection using $\mse_{\mathrm{on}}(\cdot)$. The other experimental setups are the same as in Section~\ref{sec:synthetic_setup}.

\paragraph{\textbf{Result}}
Table~\ref{tab:real} summarizes the mean and standard deviation of \textbf{Relative Regret (e)} and \textbf{Rank Correlation (e)} for the three evaluation policies ($j'=1,2,3$). Similarly to the synthetic experiment, the results demonstrate that PAS-IF improves both relative regret and rank correlation in estimator selection for every evaluation policy. Moreover, the standard deviation of PAS-IF is smaller than the non-adaptive heuristic in almost all cases, indicating the stability of the proposed method. These results provide promising empirical evidence that our PAS-IF also works fairly well in practical situations by being adaptive to an OPE problem instance.

\section{Conclusion}

We explored the problem of \textit{estimator selection for OPE}, which aims to identify the most accurate estimator among many candidate OPE estimators using only the logged data. 
Our motivation comes from the fact that the non-adaptive heuristic becomes virulent when applied to a range of evaluation policies, which is especially problematic in OPS.
With the goal of enabling a more accurate estimator selection, we proposed PAS-IF, which subsamples the logged data and imitates the importance ratio induced by the true evaluation policy, resulting in an \textit{adaptive} estimator selection. Comprehensive synthetic experiments demonstrate that PAS-IF significantly improves the accuracy of OPE and OPS compared to the non-adaptive heuristic, particularly when the evaluation policies are substantially different from the data collection policies. The real-world experiment provides additional evidence that PAS-IF enables a reliable OPE in a real bandit application. 
We hope that this work would serve as a building block for future studies of estimator selection for OPE.

% Use \bibliography{yourbibfile} instead or the References section will not appear in your paper
\bibliography{main.bbl}

\begin{thebibliography}{46}
\providecommand{\natexlab}[1]{#1}

\bibitem[{Agarwal et~al.(2017)Agarwal, Basu, Schnabel, and
  Joachims}]{agarwal2017effective}
Agarwal, A.; Basu, S.; Schnabel, T.; and Joachims, T. 2017.
\newblock {Effective Evaluation Using Logged Bandit Feedback from Multiple
  Loggers}.
\newblock \emph{KDD}, 687--696.

\bibitem[{Beygelzimer and Langford(2009)}]{beygelzimer2009offset}
Beygelzimer, A.; and Langford, J. 2009.
\newblock The Offset Tree for Learning with Partial Labels.
\newblock In \emph{Proceedings of the 15th ACM SIGKDD International Conference
  on Knowledge Discovery and Data Mining}, 129--138.

\bibitem[{Doroudi, Thomas, and Brunskill(2017)}]{doroudi2017importance}
Doroudi, S.; Thomas, P.~S.; and Brunskill, E. 2017.
\newblock Importance Sampling for Fair Policy Selection.
\newblock \emph{Grantee Submission}.

\bibitem[{Dud{\'\i}k et~al.(2014)Dud{\'\i}k, Erhan, Langford, and
  Li}]{dudik2014doubly}
Dud{\'\i}k, M.; Erhan, D.; Langford, J.; and Li, L. 2014.
\newblock Doubly Robust Policy Evaluation and Optimization.
\newblock \emph{Statistical Science}, 29(4): 485--511.

\bibitem[{Gilotte et~al.(2018)Gilotte, Calauz{\`e}nes, Nedelec, Abraham, and
  Doll{\'e}}]{gilotte2018offline}
Gilotte, A.; Calauz{\`e}nes, C.; Nedelec, T.; Abraham, A.; and Doll{\'e}, S.
  2018.
\newblock Offline A/B Testing for Recommender Systems.
\newblock In \emph{Proceedings of the 11th ACM International Conference on Web
  Search and Data Mining}, 198--206.

\bibitem[{Gruson et~al.(2019)Gruson, Chandar, Charbuillet, McInerney, Hansen,
  Tardieu, and Carterette}]{gruson2019offline}
Gruson, A.; Chandar, P.; Charbuillet, C.; McInerney, J.; Hansen, S.; Tardieu,
  D.; and Carterette, B. 2019.
\newblock Offline Evaluation to Make Decisions About Playlist Recommendation
  Algorithms.
\newblock In \emph{Proceedings of the 12th ACM International Conference on Web
  Search and Data Mining}, 420--428.

\bibitem[{Hao et~al.(2021)Hao, Ji, Duan, Lu, Szepesvari, and
  Wang}]{hao2021bootstrapping}
Hao, B.; Ji, X.; Duan, Y.; Lu, H.; Szepesvari, C.; and Wang, M. 2021.
\newblock Bootstrapping Fitted Q-Evaluation for Off-Policy Inference.
\newblock In \emph{Proceedings of the 38th International Conference on Machine
  Learning}, volume 139, 4074--4084. PMLR.

\bibitem[{Irpan et~al.(2019)Irpan, Rao, Bousmalis, Harris, Ibarz, and
  Levine}]{irpan2019off}
Irpan, A.; Rao, K.; Bousmalis, K.; Harris, C.; Ibarz, J.; and Levine, S. 2019.
\newblock Off-Policy Evaluation via Off-Policy Classification.
\newblock In \emph{Advances in Neural Information Processing Systems}.

\bibitem[{Jiang and Li(2016)}]{jiang2016doubly}
Jiang, N.; and Li, L. 2016.
\newblock Doubly Robust Off-Policy Value Evaluation for Reinforcement Learning.
\newblock In \emph{Proceedings of the 33rd International Conference on Machine
  Learning}, volume~48, 652--661. PMLR.

\bibitem[{Joachims, Swaminathan, and de~Rijke(2018)}]{joachims2018deep}
Joachims, T.; Swaminathan, A.; and de~Rijke, M. 2018.
\newblock Deep Learning with Logged Bandit Feedback.
\newblock In \emph{International Conference on Learning Representations}.

\bibitem[{Kallus, Saito, and Uehara(2021)}]{kallus2020optimal}
Kallus, N.; Saito, Y.; and Uehara, M. 2021.
\newblock Optimal Off-Policy Evaluation from Multiple Logging Policies.
\newblock In \emph{Proceedings of the 38th International Conference on Machine
  Learning}, volume 139, 5247--5256. PMLR.

\bibitem[{Kallus and Uehara(2019)}]{kallus2019intrinsically}
Kallus, N.; and Uehara, M. 2019.
\newblock Intrinsically Efficient, Stable, and Bounded Off-Policy Evaluation
  for Reinforcement Learning.
\newblock In \emph{Advances in Neural Information Processing Systems},
  volume~32.

\bibitem[{Kingma and Ba(2014)}]{kingma2014adam}
Kingma, D.~P.; and Ba, J. 2014.
\newblock Adam: A method for stochastic optimization.
\newblock \emph{arXiv preprint arXiv:1412.6980}.

\bibitem[{Kiyohara, Kawakami, and Saito(2021)}]{kiyohara2021accelerating}
Kiyohara, H.; Kawakami, K.; and Saito, Y. 2021.
\newblock Accelerating Offline Reinforcement Learning Application in Real-Time
  Bidding and Recommendation: Potential Use of Simulation.
\newblock \emph{arXiv preprint arXiv:2109.08331}.

\bibitem[{Kiyohara et~al.(2022)Kiyohara, Saito, Matsuhiro, Narita, Shimizu, and
  Yamamoto}]{kiyohara2022doubly}
Kiyohara, H.; Saito, Y.; Matsuhiro, T.; Narita, Y.; Shimizu, N.; and Yamamoto,
  Y. 2022.
\newblock Doubly Robust Off-Policy Evaluation for Ranking Policies under the
  Cascade Behavior Model.
\newblock In \emph{Proceedings of the 15th ACM International Conference on Web
  Search and Data Mining}, 487–497.

\bibitem[{Kuzborskij et~al.(2021)Kuzborskij, Vernade, Gyorgy, and
  Szepesv{\'a}ri}]{kuzborskij2021confident}
Kuzborskij, I.; Vernade, C.; Gyorgy, A.; and Szepesv{\'a}ri, C. 2021.
\newblock Confident Off-Policy Evaluation and Selection through Self-Normalized
  Importance Weighting.
\newblock In \emph{International Conference on Artificial Intelligence and
  Statistics}, 640--648. PMLR.

\bibitem[{Lepski and Spokoiny(1997)}]{lepski1997optimal}
Lepski, O.~V.; and Spokoiny, V.~G. 1997.
\newblock Optimal Pointwise Adaptive Methods in Nonparametric Estimation.
\newblock \emph{The Annals of Statistics}, 25(6): 2512--2546.

\bibitem[{Levine et~al.(2020)Levine, Kumar, Tucker, and Fu}]{levine2020offline}
Levine, S.; Kumar, A.; Tucker, G.; and Fu, J. 2020.
\newblock Offline Reinforcement Learning: Tutorial, Review, and Perspectives on
  Open Problems.
\newblock \emph{arXiv preprint arXiv:2005.01643}.

\bibitem[{Matsushima et~al.(2021)Matsushima, Furuta, Matsuo, Nachum, and
  Gu}]{matsushima2021deployment}
Matsushima, T.; Furuta, H.; Matsuo, Y.; Nachum, O.; and Gu, S. 2021.
\newblock Deployment-Efficient Reinforcement Learning via Model-Based Offline
  Optimization.
\newblock In \emph{International Conference on Learning Representations}.

\bibitem[{Metelli, Russo, and Restelli(2021)}]{metelli2021subgaussian}
Metelli, A.~M.; Russo, A.; and Restelli, M. 2021.
\newblock Subgaussian and Differentiable Importance Sampling for Off-Policy
  Evaluation and Learning.
\newblock 34.

\bibitem[{Narita, Yasui, and Yata(2021)}]{narita2021debiased}
Narita, Y.; Yasui, S.; and Yata, K. 2021.
\newblock Debiased Off-Policy Evaluation for Recommendation Systems.
\newblock In \emph{Proceedings of 15th ACM Conference on Recommender Systems},
  372--379.

\bibitem[{Paine et~al.(2020)Paine, Paduraru, Michi, Gulcehre, Zolna, Novikov,
  Wang, and de~Freitas}]{paine2020hyperparameter}
Paine, T.~L.; Paduraru, C.; Michi, A.; Gulcehre, C.; Zolna, K.; Novikov, A.;
  Wang, Z.; and de~Freitas, N. 2020.
\newblock Hyperparameter Selection for Offline Reinforcement Learning.
\newblock \emph{arXiv preprint arXiv:2007.09055}.

\bibitem[{Paszke et~al.(2019)Paszke, Gross, Massa, Lerer, Bradbury, Chanan,
  Killeen, Lin, Gimelshein, Antiga et~al.}]{paszke2019pytorch}
Paszke, A.; Gross, S.; Massa, F.; Lerer, A.; Bradbury, J.; Chanan, G.; Killeen,
  T.; Lin, Z.; Gimelshein, N.; Antiga, L.; et~al. 2019.
\newblock Pytorch: An Imperative Style, High-Performance Deep Learning Library.
\newblock \emph{Advances in Neural Information Processing Systems}, 32:
  8026--8037.

\bibitem[{Pedregosa et~al.(2011)Pedregosa, Varoquaux, Gramfort, Michel,
  Thirion, Grisel, Blondel, Prettenhofer, Weiss, Dubourg
  et~al.}]{pedregosa2011scikit}
Pedregosa, F.; Varoquaux, G.; Gramfort, A.; Michel, V.; Thirion, B.; Grisel,
  O.; Blondel, M.; Prettenhofer, P.; Weiss, R.; Dubourg, V.; et~al. 2011.
\newblock Scikit-learn: Machine Learning in Python.
\newblock \emph{Journal of Machine Learning Research}, 12: 2825--2830.

\bibitem[{Precup, Sutton, and Singh(2000)}]{precup2000eligibility}
Precup, D.; Sutton, R.~S.; and Singh, S.~P. 2000.
\newblock Eligibility Traces for Off-Policy Policy Evaluation.
\newblock In \emph{Proceedings of the 17th International Conference on Machine
  Learning}, 759–766.

\bibitem[{Saito et~al.(2021{\natexlab{a}})Saito, Aihara, Matsutani, and
  Narita}]{saito2020open}
Saito, Y.; Aihara, S.; Matsutani, M.; and Narita, Y. 2021{\natexlab{a}}.
\newblock Open Bandit Dataset and Pipeline: Towards Realistic and Reproducible
  Off-Policy Evaluation.
\newblock In \emph{Thirty-fifth Conference on Neural Information Processing
  Systems Datasets and Benchmarks Track}.

\bibitem[{Saito and Joachims(2021)}]{saito2021counterfactual}
Saito, Y.; and Joachims, T. 2021.
\newblock Counterfactual Learning and Evaluation for Recommender Systems:
  Foundations, Implementations, and Recent Advances.
\newblock In \emph{Proceedings of the 15th ACM Conference on Recommender
  Systems}, 828–830.

\bibitem[{Saito and Joachims(2022)}]{saito2022off}
Saito, Y.; and Joachims, T. 2022.
\newblock Off-Policy Evaluation for Large Action Spaces via Embeddings.
\newblock In \emph{International Conference on Machine Learning}, 19089--19122.
  PMLR.

\bibitem[{Saito et~al.(2021{\natexlab{b}})Saito, Udagawa, Kiyohara, Mogi,
  Narita, and Tateno}]{saito2021evaluating}
Saito, Y.; Udagawa, T.; Kiyohara, H.; Mogi, K.; Narita, Y.; and Tateno, K.
  2021{\natexlab{b}}.
\newblock Evaluating the Robustness of Off-Policy Evaluation.
\newblock In \emph{Proceedings of the 15th ACM Conference on Recommender
  Systems}, 114–123.

\bibitem[{Saito, Udagawa, and Tateno(2021)}]{saito2021data}
Saito, Y.; Udagawa, T.; and Tateno, K. 2021.
\newblock Data-Driven Off-Policy Estimator Selection: An Application in User
  Marketing on An Online Content Delivery Service.
\newblock \emph{arXiv preprint arXiv:2109.08621}.

\bibitem[{Strehl et~al.(2010)Strehl, Langford, Li, and
  Kakade}]{strehl2010learning}
Strehl, A.; Langford, J.; Li, L.; and Kakade, S.~M. 2010.
\newblock Learning from Logged Implicit Exploration Data.
\newblock In \emph{Advances in Neural Information Processing Systems},
  volume~23, 2217--2225.

\bibitem[{Su et~al.(2020)Su, Dimakopoulou, Krishnamurthy, and
  Dud{\'\i}k}]{su2020doubly}
Su, Y.; Dimakopoulou, M.; Krishnamurthy, A.; and Dud{\'\i}k, M. 2020.
\newblock Doubly Robust Off-Policy Evaluation with Shrinkage.
\newblock In \emph{Proceedings of the 37th International Conference on Machine
  Learning}, volume 119, 9167--9176. PMLR.

\bibitem[{Su, Srinath, and Krishnamurthy(2020)}]{su2020adaptive}
Su, Y.; Srinath, P.; and Krishnamurthy, A. 2020.
\newblock Adaptive Estimator Selection for Off-Policy Evaluation.
\newblock \emph{arXiv preprint arXiv:2002.07729}.

\bibitem[{Su et~al.(2019)Su, Wang, Santacatterina, and Joachims}]{su2019cab}
Su, Y.; Wang, L.; Santacatterina, M.; and Joachims, T. 2019.
\newblock CAB: Continuous Adaptive Blending for Policy Evaluation and Learning.
\newblock In \emph{International Conference on Machine Learning}, volume~84,
  6005--6014.

\bibitem[{Swaminathan and Joachims(2015{\natexlab{a}})}]{swaminathan2015batch}
Swaminathan, A.; and Joachims, T. 2015{\natexlab{a}}.
\newblock {Batch Learning from Logged Bandit Feedback through Counterfactual
  Risk Minimization}.
\newblock \emph{Journal of Machine Learning Research}, 16: 1731--1755.

\bibitem[{Swaminathan and Joachims(2015{\natexlab{b}})}]{swaminathan2015self}
Swaminathan, A.; and Joachims, T. 2015{\natexlab{b}}.
\newblock The Self-Normalized Estimator for Counterfactual Learning.
\newblock In \emph{Advances in Neural Information Processing Systems},
  volume~28, 3231--3239.

\bibitem[{Tang and Wiens(2021)}]{tang2021model}
Tang, S.; and Wiens, J. 2021.
\newblock Model Selection for Offline Reinforcement Learning: Practical
  Considerations for Healthcare Settings.
\newblock In \emph{Machine Learning for Healthcare Conference}, 2--35. PMLR.

\bibitem[{Thomas and Brunskill(2016)}]{thomas2016data}
Thomas, P.; and Brunskill, E. 2016.
\newblock Data-Efficient Off-Policy Policy Evaluation for Reinforcement
  Learning.
\newblock In \emph{Proceedings of the 33rd International Conference on Machine
  Learning}, volume~48, 2139--2148. PMLR.

\bibitem[{Thomas, Theocharous, and
  Ghavamzadeh(2015{\natexlab{a}})}]{thomas2015confidence}
Thomas, P.; Theocharous, G.; and Ghavamzadeh, M. 2015{\natexlab{a}}.
\newblock High Confidence Policy Improvement.
\newblock In \emph{Proceedings of the 32th International Conference on Machine
  Learning}, 2380--2388.

\bibitem[{Thomas, Theocharous, and
  Ghavamzadeh(2015{\natexlab{b}})}]{thomas2015high}
Thomas, P.~S.; Theocharous, G.; and Ghavamzadeh, M. 2015{\natexlab{b}}.
\newblock {High-Confidence Off-Policy Evaluation}.
\newblock 3000--3006.

\bibitem[{Tucker and Lee(2021)}]{tucker2021improved}
Tucker, G.; and Lee, J. 2021.
\newblock Improved Estimator Selection for Off-Policy Evaluation.

\bibitem[{Voloshin et~al.(2019)Voloshin, Le, Jiang, and
  Yue}]{voloshin2019empirical}
Voloshin, C.; Le, H.~M.; Jiang, N.; and Yue, Y. 2019.
\newblock Empirical Study of Off-Policy Policy Evaluation for Reinforcement
  Learning.
\newblock \emph{arXiv preprint arXiv:1911.06854}.

\bibitem[{Wang, Agarwal, and Dudik(2017)}]{wang2017optimal}
Wang, Y.-X.; Agarwal, A.; and Dudik, M. 2017.
\newblock {Optimal and Adaptive Off-policy Evaluation in Contextual Bandits}.
\newblock In \emph{Proceedings of the 34th International Conference on Machine
  Learning}, 3589--3597.

\bibitem[{Yang et~al.(2021)Yang, Qi, Cui, and Chen}]{yang2021pessimistic}
Yang, C.-H.~H.; Qi, Z.; Cui, Y.; and Chen, P.-Y. 2021.
\newblock Pessimistic Model Selection for Offline Deep Reinforcement Learning.
\newblock \emph{arXiv preprint arXiv:2111.14346}.

\bibitem[{Yang et~al.(2020)Yang, Dai, Nachum, Tucker, and
  Schuurmans}]{yang2020offline}
Yang, M.; Dai, B.; Nachum, O.; Tucker, G.; and Schuurmans, D. 2020.
\newblock Offline Policy Selection under Uncertainty.
\newblock \emph{arXiv preprint arXiv:2012.06919}.

\bibitem[{Zhang and Jiang(2021)}]{zhang2021towards}
Zhang, S.; and Jiang, N. 2021.
\newblock Towards Hyperparameter-free Policy Selection for Offline
  Reinforcement Learning.
\newblock In \emph{Advances in Neural Information Processing Systems},
  volume~34.

\end{thebibliography}

\clearpage
\appendix
\clearpage
\appendix

\section{Gradient Derivation} \label{app:deviation}
\subsection{Importance Fitting Objective}
Here, we derive the following gradient and show its direction.
\begin{align*}
    \frac{\partial d}{\partial \rho_{\theta}} 
    &\propto
    \frac{(\tw - w)}{(1-\rho_{\theta})^2} \left( \frac{1}{\mE_{\pi_b}[\rho_{\theta}]} - 1 \right) \\
    & \quad \cdot \left( 1 - \frac{\pi_b\rho_{\theta} (1 - \rho_{\theta})}{(\mE_{\pi_b}[\rho_{\theta}])(1 - \mE_{\pi_b}[\rho_{\theta}])} \right).
\end{align*}
To start, we decompose the pseudo importance ratio as follows.
\begin{align*}
    & \tw(x, a) \\
    & = \frac{\tpi_e(a|x)}{\tpi_b(a|x)} \\
    &= \left( \pi_b(a|x) \frac{\rho_{\theta}(x,a)}{\mE_{\pi_b(a|x)[\rho_{\theta}(a|x)]}} \right) \\
    & \quad \cdot \left( \pi_b(a|x) \frac{1 - \rho_{\theta}(x,a)}{1 - \mE_{\pi_b(a|x)[\rho_{\theta}(a|x)]}} \right)^{-1} \\
    &= \rho_{\theta}(x,a) (1 - \rho_{\theta}(x,a))^{-1} \\
    & \quad \cdot \left(\mE_{\pi_b(a|x)}[\rho_{\theta}(x,a)]\right)^{-1} \left(1-\mE_{\pi_b(a|x)}[\rho_{\theta}(x,a)]\right)^{-1} \\
    &= \underbrace{\rho_{\theta}(x,a)}_{A} \underbrace{(1 - \rho_{\theta}(x,a))^{-1}}_{B} \\
    & \quad \cdot \underbrace{\left(\sum_{a \in \calA}\pi_b(a|x) \rho_{\theta}(x,a)\right)^{-1}}_{C} \underbrace{\left(1-\sum_{a \in \calA}\pi_b(a|x)\rho_{\theta}(x,a)\right)}_{D}.
\end{align*}
Then, we calculate the gradient of $d(w,\tw) \left(= (w(x,a)-\tw(x,a))^2 \right)$ for any given $(x, a)$ as follows.
\begin{align*}
    & \frac{\partial d}{\partial \rho_{\theta}} \\
    &= (2\tw - 2w) \cdot \frac{\partial \tw}{\partial \rho_{\theta}} \\
    & \propto (\tw - w) \\
    & \quad \cdot \left( BCD - B^2 \cdot (-1) ACD - C^2 \cdot \pi_b ABD - \pi_b ABC \right) \\
    &= (\tw - w) \cdot BC \cdot \left( D (1 + AB) - \pi_b A (1 + CD) \right) \\
    &= (\tw - w) \cdot BC \\
    & \quad \cdot \left( D \left( 1 + \frac{\rho_{\theta}}{1-\rho} \right) - \pi_b A \left(1 + \frac{1 - \mE_{\pi_b}[\rho_{\theta}]}{\mE_{\pi_b}[\rho_{\theta}]} \right) \right) \\
    &= (\tw - w) \cdot BC \cdot \left( D \left( \frac{1}{1-\rho_{\theta}} \right) - \pi_b A \left(\frac{1}{\mE_{\pi_b}[\rho_{\theta}]} \right) \right) \\
    &= \frac{(\tw - w)}{(1-\rho_{\theta}) \mE_{\pi_b}[\rho_{\theta}]} \left( \frac{1 - \mE_{\pi_b}[\rho_{\theta}]}{1-\rho_{\theta}} - \pi_b \frac{\rho_{\theta}}{\mE_{\pi_b}[\rho_{\theta}]} \right) \\
    &= \frac{(\tw - w)}{(1-\rho_{\theta})^2} \frac{1 - \mE_{\pi_b}[\rho_{\theta}]}{\mE_{\pi_b}[\rho_{\theta}]} \left( 1 - \pi_b \frac{\rho_{\theta} (1 - \rho_{\theta})}{(\mE_{\pi_b}[\rho_{\theta}])(1 - \mE_{\pi_b}[\rho_{\theta}])} \right) \\
    &= \frac{(\tw - w)}{(1-\rho_{\theta})^2} \left( \frac{1}{\mE_{\pi_b}[\rho_{\theta}]} - 1 \right) \left( 1 - \frac{\pi_b \rho_{\theta} (1 - \rho_{\theta})}{(\mE_{\pi_b}[\rho_{\theta}])(1 - \mE_{\pi_b}[\rho_{\theta}])} \right).
\end{align*}
The gradient direction is determined by $(\tw - w)$, as the second and the third terms are always non-negative. In particular, we have the following for the third term.
\begin{enumerate}
    \item If $\mE_n[\rho] < \rho$,
    \begin{align*}
        \frac{\pi_b \rho_{\theta} (1 - \rho_{\theta})}{(\mE_{\pi_b}[\rho_{\theta}])(1 - \mE_{\pi_b}[\rho_{\theta}])} 
        \leq \frac{\pi_b \rho_{\theta}}{\mE_{\pi_b}[\rho_{\theta}]}
        = \frac{\pi_b \rho_{\theta}}{\sum_{a \in \calA}\pi_b \rho_{\theta}}
        \leq 1.
    \end{align*}
    \item Otherwise,
    \begin{align*}
        &\frac{\pi_b \rho_{\theta} (1 - \rho_{\theta})}{(\mE_{\pi_b}[\rho_{\theta}])(1 - \mE_{\pi_b}[\rho_{\theta}])}
        \leq \frac{\pi_b (1-\rho_{\theta})}{1-\mE_{\pi_b}[\rho_{\theta}]} \\
        &= \frac{\pi_b (1-\rho_{\theta})}{\sum_{a \in \calA}\pi_b (1 -\rho_{\theta})} 
        \leq 1.
    \end{align*}
\end{enumerate}

\subsection{Regularization}
We also derive the gradient of $R(\tpi, k) := \mE_{p(x)}[( \mE_{\pi_b(a|x)}[ \rho_{\theta}(x,a)] - k )^2]$ as follows.
\begin{align*}
    \frac{\partial R}{\partial \theta} 
    &= \mE_{p(x)} \bigg[ 2 \underbrace{\left( \mE_{\pi_b(a|x)}[ \rho_{\theta} ] - k \right)}_{\text{independent of } \pi_b(\cdot|x)} \mE_{\pi_b(a|x)}\left[ \frac{\partial \rho_{\theta}}{\partial \theta} \right] \bigg] \\
    & \propto \mE_{p(x)\pi_b(a|x)} \left[ \left(\mE_{\pi_b(a|x)}[\rho_{\theta}] - k \right) \frac{\partial \rho_{\theta}}{\partial \theta} \right]. 
\end{align*}

\section{Additional Benefit of PAS-IF} \label{app:additional_benefit}
The main text assumed that the behavior policy consists of $l \, (\geq 1)$ different data collection policies as follows.
\begin{align*}
  \pi_b(a|x) := \sum_{\pi_j \in \{ \pi_1, \cdots, \pi_l\}} p(j) \pi_j(a|x),
\end{align*}
where $p(j)$ is a \textit{meta} policy assignment rule, which includes situations such as A/B tests.
When we apply the non-adaptive heuristic, practitioners have to deploy at least two policies ($l \geq 2$) in the A/B test, as one of them must be used as the pseudo evaluation policy. This is one of the drawbacks of the non-adaptive heuristic, because additionally implementing new policies aside from the one used in daily operation might be costly and harmful. On the other hand, our PAS-IF is applicable even when there is only a single data collection policy ($l = 1$), which is naturally obtained through platform operations.

Moreover, PAS-IF can be applied to the following more general class of behavior policies (where the non-adaptive heuristic cannot be used).
\begin{align*}
  \pi_b(a|x) := \sum_{\pi_j \in \{ \pi_1, \cdots, \pi_l\}} \red{p(j | x)} \pi_j(a|x),
\end{align*}
where $p(j|x)$ is a \textit{context-dependent} meta policy assignment rule, which is either stochastic or deterministic. This enables us to use more diverse and large-scale data collected via platform operations. For example, we sometimes assign different policies depending on time period, season, or user's demographic profile. PAS-IF is able to leverage all these kinds of general logged data where the non-adaptive heuristic is not applicable.

\section{End-to-End OPE Procedure with PAS-IF}
Algorithm~\ref{algo:whole} describes an end-to-end OPE procedure, including our PAS-IF as a subroutine for estimator selection.
In Algorithm~\ref{algo:whole}, we perform OPE and estimator selection on different subsets of the logged data to avoid a potential bias~\citep{thomas2015confidence}. 
Specifically, Algorithm~\ref{algo:whole} first splits $\calD_b$ into $\calD_{b}^{(pre)}$ and $\calD_b^{(post)}$, and then uses $\calD_{b}^{(pre)}$ only for estimator selection and $\calD_b^{(post)}$ for OPE.
Algorithm~\ref{algo:whole} iterates this procedure with different data splits and use the averaged estimate as the final OPE result.

\begin{figure}[tb]
\centering
\begin{algorithm}[H]
\caption{OPE with PAS-IF}
\begin{algorithmic}[1]
\REQUIRE a candidate set of OPE estimators $\mathcal{V}$, logged bandit dataset $\calD_b$, evaluation policy $\pi_e$, sets of random seeds $\calS_1, \calS_2$
\ENSURE estimated policy value $\hat{V}(\pi_e; \calD_b)$
\FOR{$s_1 \in \calS_1$}
    \STATE $ \calD_{b}^{(pre)}, \calD_b^{(post)} \leftarrow \mathrm{UniformSubsample}(\calD_b; s_1) $
    \STATE $ \hat{V} \leftarrow \textrm{PAS-IF}(\mathcal{V}; \calD_{b}^{(pre)}, \pi_e, \calS_2) $ $\quad \triangleright \, \mathrm{Algorithm 1}$
    \STATE $ z_{s_1} \leftarrow \hat{V}(\pi_e; \calD_b^{(post)}) $
\ENDFOR
\STATE $ \hat{V}(\pi_e; \calD_b) \leftarrow \sum_{s_1 \in \calS_1} z_{s_1} / |\calS_1| $
\end{algorithmic}
\label{algo:whole}
\end{algorithm}
\end{figure}

\section{Off-Policy Evaluation Details} \label{app:estimators}
Here, we summarize the definition and the important statistical properties of OPE estimators used in the experiment. We also describe how to train $\hat{q}$ by cross-fitting~\citep{narita2021debiased} and tune the estimators' built-in hyperparameters using SLOPE~\citep{su2020adaptive, tucker2021improved}.

\subsection{Off-Policy Estimators}

\paragraph{Direct Method (DM).}
DM~\citep{beygelzimer2009offset} first trains a supervised machine learning method, such as logistic regression, to estimate the mean reward function $q(x,a) := \mE[r|x,a]$. Then, DM estimates the policy value as follows.
\begin{align*}
    \dm := \frac{1}{n} \sum_{i=1}^n \sum_{a \in \calA} \pi_e(a|x_i) \hat{q}(x_i, a),
\end{align*}
where $\hat{q}(x_i, a_i)$ is the predicted reward. DM performs reasonably well when the reward prediction is accurate. However, DM incurs high bias when the reward prediction fails.

\paragraph{Inverse Propensity Scoring (IPS).} 
Rather than performing a reward regression, IPS~\citep{precup2000eligibility, strehl2010learning} uses importance sampling and addresses the distribution shift between the behavior and evaluation policies as follows.
\begin{align*}
    \ips := \frac{1}{n} \sum_{i=1}^n w(x_i, a_i) r_i,
\end{align*}
where $w(x_i, a_i) := \pi_e(a_i|x_i) / \pi_b(a_i|x_i)$ is the importance ratio. IPS is unbiased under some identification assumptions such as full support ($\pi_e(a|x) > 0 \rightarrow \pi_b(a|x), \forall (x,a) \in \calX \times \calA$). However, when the behavior and evaluation policies are dissimilar or the action space is large, IPS often suffers from high variance~\citep{saito2022off}.

\paragraph{Doubly Robust (DR).}
DR~\citep{thomas2016data, jiang2016doubly, dudik2014doubly} aims to reduce the variance of IPS by combining DM and IPS. Specifically, DR uses the predicted reward as a control variate and applies importance sampling only to its residual as follows.
\begin{align*}
    & \dr \\ 
    &:= \frac{1}{n} \sum_{i=1}^n \left( w(x_i, a_i) (r_i - q(x_i, a_i)) + \sum_{a \in \calA} \pi_e(a|x_i) \hat{q}(x_i, a) \right).
\end{align*}
When $\hat{q}$ is reasonably accurate, DR reduces the variance of IPS, while remaining unbiased. However, DR can still suffer from high variance when the action space is large~\citep{saito2022off}.

\paragraph{Clipping estimators (IPSps, DRps).} 
One possible approach to alleviate the variance of DR and IPS is to clip the large importance ratio as follows.
\begin{align*}
    w_{\mathrm{ps}}(x_i,a_i) := \min \{ w(x_i,a_i), \lambda \},
\end{align*}
where $\lambda \geq 0$ is the clipping threshold.
When using the clipped importance ratio, IPS and DR are no longer unbiased, but we can reduce their variance by eliminating the large importance ratio. We call the resulting estimators as \textbf{P}essimistic \textbf{S}hrinkage estimators (IPSps and DRps) following~\citet{su2020doubly}.

\paragraph{Self-Normalized estimators (SNIPS, SNDR).} 
The Self-Normalized estimators~\citep{swaminathan2015self,joachims2018deep} normalize the importance ratio to gain some stability as follows.
\begin{align*}
    w_{\mathrm{norm}}(x_i,a_i) := \frac{w(x_i,a_i)}{\sum_{j=1}^n w(x_j,a_j) / n}. 
\end{align*}
SN estimators is biased, but is still consistent. Moreover, it is known to sometimes greatly reduce the variance of IPS~\citep{kallus2019intrinsically}.

\paragraph{Switch.}
Switch~\citep{wang2017optimal} aims to reduce the variance of DR by interpolating between DM and DR. Specifically, Switch uses DR only when the importance ratio is not large. Otherwise, it uses DM, which has a low variance. Specifically, this estimator alternatively defines the importance ratio as follows.
\begin{align*}
    w_{\mathrm{switch}}(x_i, a_i) := w(x_i,a_i) \mathbb{I} \{ w(x_i,a_i) \leq \lambda \},
\end{align*}
where $\lambda \geq 0$ is the threshold hyperparameter. When $\lambda = 0$, Switch is identical to DM, while $\lambda \rightarrow \infty$ yields DR.

\paragraph{DR with Optimistic Shrinkage (DRos).} DRos~\citep{su2020doubly} bases on a motivation of reducing the variance of DR, which is similar to that of Switch, but does so by directly minimizing the sharp bound of MSE by applying the following shrunk version of the importance ratio.
\begin{align*}
    w_{\mathrm{os}}(x_i,a_i) := \frac{\lambda }{w(x_i,a_i)^2 + \lambda} w(x_i,a_i),
\end{align*}
where $\lambda \geq 0$ is a parameter that controls the bias and variance. When $\lambda = 0$, $w_{\mathrm{os}}(x_i,a_i) = 0$ leading to DM.
In contrast, when $\lambda \rightarrow \infty$, $w_{\mathrm{os}}(x_i,a_i) = w(x_i, a_i)$ leading to DR.

\paragraph{Subgaussian importance sampling estimators (IPS-$\lambda$, DR-$\lambda$).}
To improve the error concentration rate from the exponential rate to subgaussian rate,
$\lambda$-estimators~\citep{metelli2021subgaussian} apply smooth shrinkage to the importance ratio as follows. 
\begin{align*}
    w_{\lambda}(x_i,a_i) := ((1 - \lambda) w(x_i,a_i)^s + \lambda)^{\frac{1}{s}}
\end{align*}
where $\lambda \in [0, 1]$ and $s \in (-\infty,1]$ are the hyperprameters. In our experiments, we use $s=-1$, which is the main proposal of \citet{metelli2021subgaussian}. When $\lambda=0$, $w_{\lambda}(x_i,a_i) = w(x_i,a_i)$ leading to DR, whereas $\lambda=1$ results in $w_{\lambda}(x_i,a_i) = 1$. As $\lambda$ becomes large, the resulting estimators reduce their variance, while increase their bias.

\begin{figure*}[h]
  \centering
  \begin{minipage}[b]{0.40\linewidth}
  \centering
  \includegraphics[width=0.9\linewidth]{figs/label.png}
  \vspace{1mm}
  \end{minipage} \\
  \begin{minipage}[b]{0.85\linewidth}
  \centering
      \begin{minipage}[b]{0.45\linewidth}
        \centering
        \includegraphics[width=1.05\linewidth]{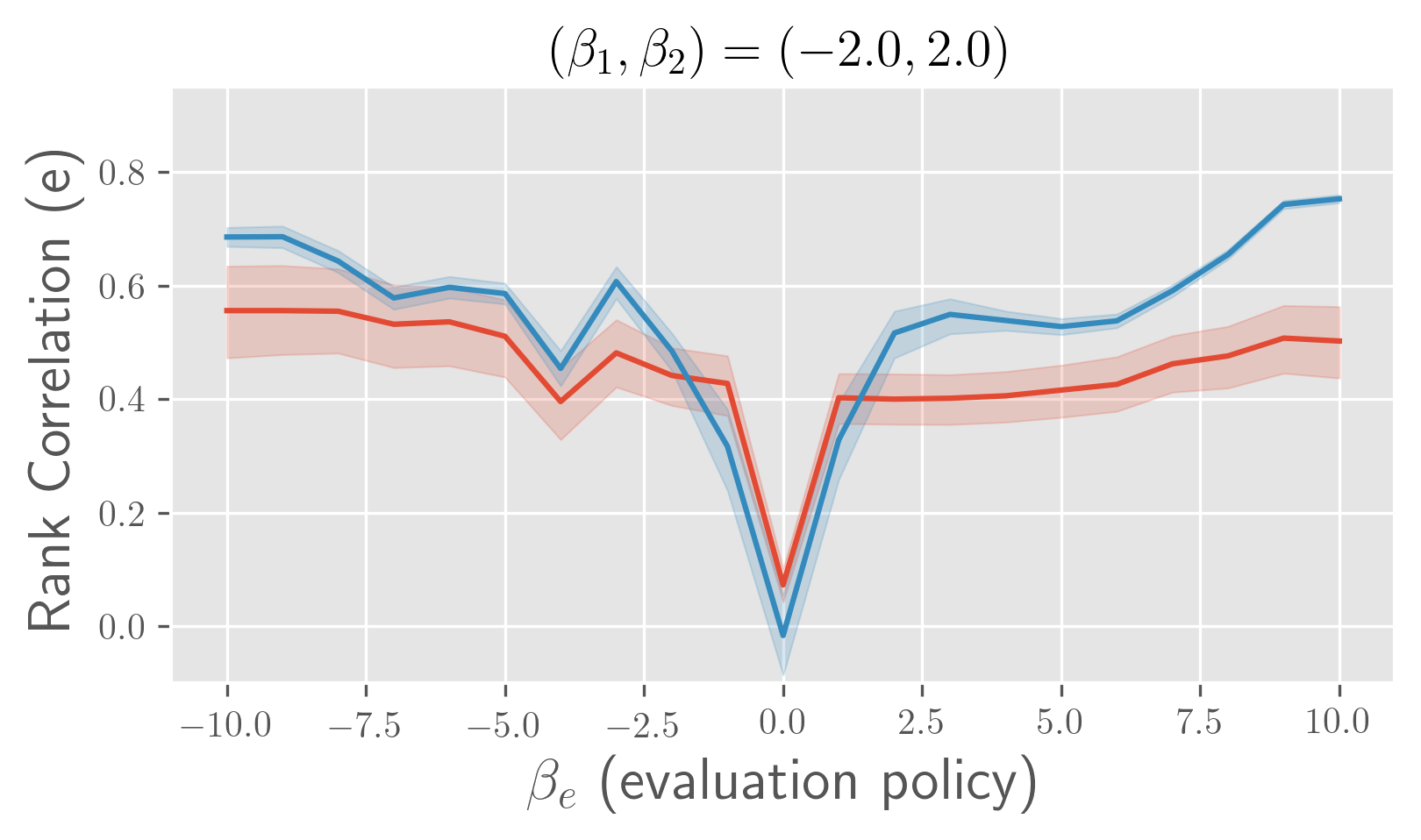}
      \end{minipage}
      \begin{minipage}[b]{0.45\linewidth}
        \centering
        \includegraphics[width=1.05\linewidth]{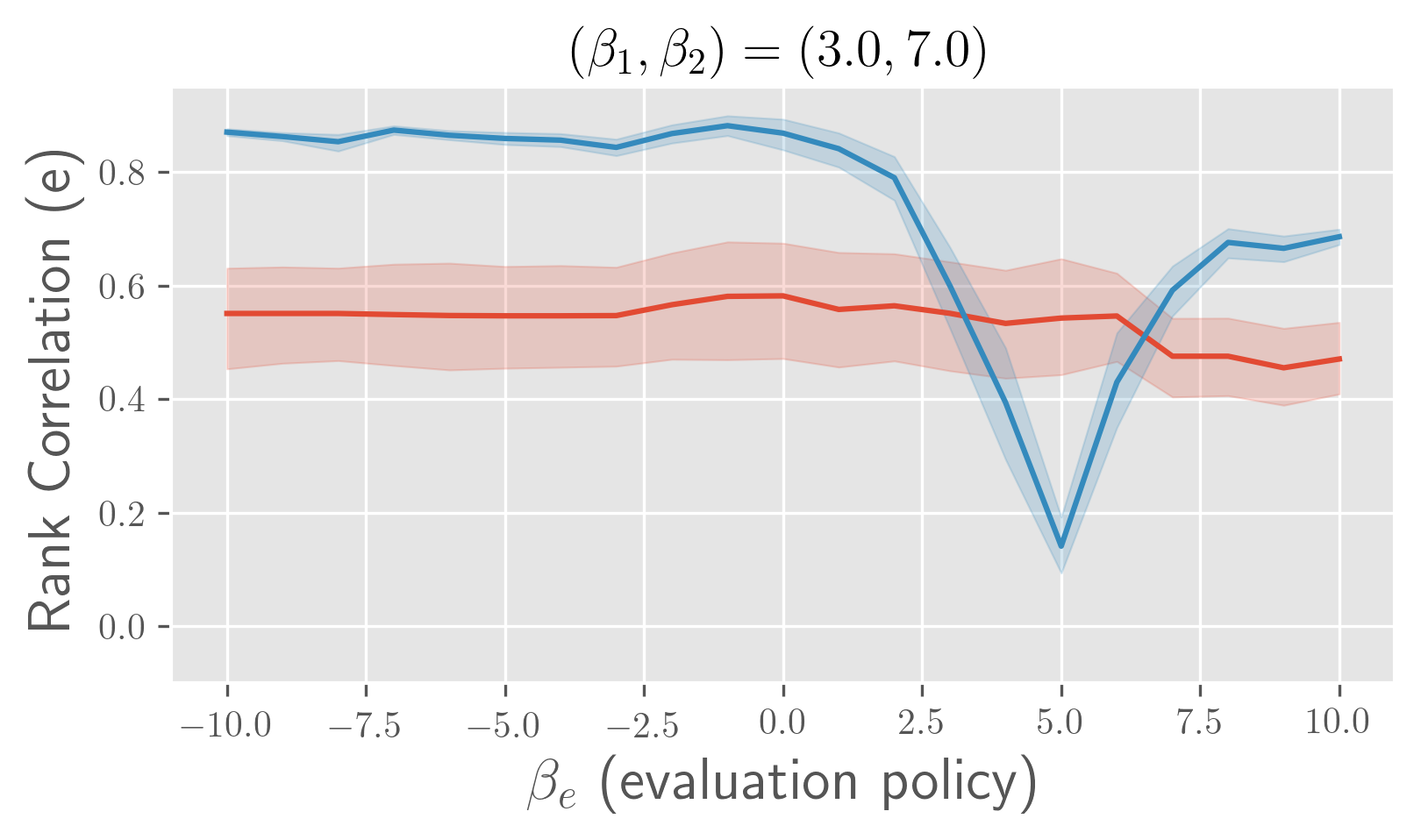}
      \end{minipage}
      \vspace{-1mm}
      \caption{\textbf{Rank Correlation (e)} with varying evaluation policies ($\beta_e$) in estimator selection in the synthetic experiment.}
      \label{fig:rankcorr}
  \end{minipage}
\end{figure*}

\subsection{Related Techniques}

\paragraph{Cross-fitting procedure.}
To avoid the potential bias caused by overfitting of the reward predictor $\hat{q}$, we train it and construct DM and DR-based estimators via the following cross-fitting procedure~\citep{narita2021debiased}:
\begin{enumerate}
    \item Given size $n$ of logged data $\calD$, take a $K$-fold random partition $(\calD_{\kappa})_{\kappa=1}^K$, each of which contains $n_{\kappa} = n / K$ samples. We also define $\calD_{\kappa}^c := \calD \setminus \calD_{\kappa}$.
    \item For each $\kappa=1,\ldots,K$, construct reward predictors $\{ \hat{q}_{\kappa} \}_{\kappa=1}^K$ using the subset of data $\calD_{\kappa}^c$.
    \item Estimate the policy value by $K^{-1} \sum_{\kappa=1}^K \hat{V}(\pi_e; \calD_k, \hat{q}_k)$.
\end{enumerate}
where we use $K=3$ in all our experiments.

\paragraph{Hyperparameter tuning of OPE via SLOPE.} To tune the estimators' built-in tradeoff hyperparameter $\lambda$, we use the SLOPE procedure~\citep{su2020adaptive,tucker2021improved}. SLOPE is applicable when the candidate set of hyperparameters $\Lambda := \{ \lambda_m \}_{m=1}^M$ satisfy the following \textit{monotonicity} assumption~\citep{tucker2021improved}. 
\begin{enumerate}
    \item $\mathrm{Bias}(\hat{V}(\cdot; \lambda_m)) \leq \mathrm{Bias}(\hat{V}(\cdot; \lambda_{m+1})), \forall m \in [M - 1]$
    \item $\mathrm{CNF}(\hat{V}(\cdot; \lambda_m)) \geq \mathrm{CNF}(\hat{V}(\cdot; \lambda_{m+1})), \forall m \in [M - 1]$
\end{enumerate}
where $\mathrm{CNF}(\cdot)$ is a high probability bound on the deviation of $\hat{V}$ such as Hoeffding and Bernstein bounds.
Then, SLOPE selects the hyperparameter as follows.
\begin{align*}
    \hat{m} 
    &:= \max \{m \in [M] : |\hat{V}(\cdot; \lambda_m) - \hat{V}(\cdot; \lambda_{m'})| \\
    & \quad \leq \mathrm{CNF}(\hat{V}(\cdot; \lambda_m)) + (\sqrt{6} - 1) \mathrm{CNF}(\hat{V}(\cdot; \lambda_{m'})), \\
    & \quad \forall m' < m \}
\end{align*}
As for the theoretical guarantee of SLOPE, we refer the reader to \citet{tucker2021improved}.

\section{Additional Experimental Details} \label{app:experiment_detail}
This section describes the experiment details omitted in the main text.

\paragraph{Candidate policies in the OPS task.}
In total, we prepare 20 policies in the OPS task. We use the OPL methods provided by OBP. In OBP, each OPL method is specified by the combination of learning algorithms and base models. The learning methods are categorized into IPWLearner and QLearner. IPWLearner solves a cost-sensitive multi-class classification problem to maximize the policy value estimated by IPS. QLearner estimates the reward predictor $\hat{q}$ to specify the most promising actions. As the base models for each method, we use LogisticRegression and RandomForest provided by scikit-learn~\citep{pedregosa2011scikit}. After obtaining the logit values for each $(x, a)$ pair in classification or $\hat{q}(x, a)$ in the reward prediction, we define candidate policies using Eq.~\eqref{eq:policy}. We vary the inverse temperature parameter of Eq.~\eqref{eq:policy} as $\beta = \{ 1, 2, 10, 20, 100 \}$.

\paragraph{Estimators' built-in hyperparameters.} 
For IPSps, DRps, Switch, and DRos, we set $\Lambda = \{ 100, 500, 1000, 5000, \ldots, 10^5, \infty \}$. For IPS-$\lambda$ and DR-$\lambda$, we set $\Lambda = \{ 0.1, 0.2, \ldots, 1.0 \}$. We pick $\lambda$ from the above candidate set ($\Lambda$) via SLOPE~\citep{su2020adaptive, tucker2021improved} before applying estimator selection by PAS-IF or non-adaptive heuristic.

\paragraph{Ground-truth MSE.} 
The evaluation metrics (e) require the ground-truth MSE of each OPE estimator. We first calculate $V(\pi_e)$ on 1,000,000 samples which are independent of $\calD_b$. We then derive $\hat{V}(\pi_e; \calD_b^{\prime})$ on 100 test logged datasets, which are also independent of $\calD_b$.
Finally, we calculate the (true) MSE using $V(\pi_e)$ and 100 realizations of $\hat{V}(\pi_e; \calD_b^{\prime})$.

\paragraph{Neural subsampling function.}
We parametrize $\rho_{\theta}$ using a neural network $f_{\theta}$ as follows.
\begin{align*}
    \rho_{\theta}(x, a) := \mathrm{sigmoid}(f_{\theta}(x,a)),
\end{align*}
where $\mathrm{sigmoid}(z) := (1 + \exp(-z))^{-1}$ maps the output of $f_{\theta}(\cdot)$ into the range of $(0,1)$. $f_{\theta}$ consists of a 3-layer MLP with 100 dimensional hidden layers and ReLU activation. We implement $f_{\theta}$ with PyTorch~\citep{paszke2019pytorch} and use Adam~\cite{kingma2014adam} as its optimizer.

\begin{table}[htb]
\large
\centering
\caption{\textbf{Relative Regret (p)} and \textbf{Rank Correlation (p)} in OPS in the synthetic experiment: $(\beta_1,\beta_2)=(3.0,7.0)$.} \label{tab:ops3-7}
\def\arraystretch{1.2}
\scalebox{0.7}{
\begin{tabular}{c|cc}
\toprule
 & \textbf{Relative Regret (p)} & 
\textbf{Rank Correlation (p)} \\\midrule \midrule
non-adaptive heuristic
& 0.0325 \, ($\pm$ 0.0430)
& 0.778 \, ($\pm$ 0.491)
\\
PAS-IF (ours)
& \textbf{0.0230} \, ($\pm$ 0.0225)
& \textbf{0.972} \, ($\pm$ 0.0191)
\\
\bottomrule
\end{tabular}
}
\vskip 0.1in
\raggedright
\fontsize{9pt}{9pt}\selectfont \textit{Note}:
A lower value is better for Relative Regret (p), while a higher value is better for Rank Correlation (p). 
The \textbf{bold} fonts represent the best estimator selection method. The values in parantheses indicate standard deviation of the metrics.
\end{table}

\paragraph{Rank Correlation (e) in estimator selection.}
Figure~\ref{fig:rankcorr} shows the mean and 95\% confidence intervals of the rank correlation in the estimator selection task. The results indicate that PAS-IF improves the rank correlation more as the divergence between the behavior and evaluation policies grows, suggesting the further argument for the effectiveness of our adaptive method. On the other hand, we also observe that the rank correlation of the non-adaptive heuristics and PAS-IF drops shapely around the cases where $\beta_e \simeq (\beta_1 + \beta_2) / 2$. However, this observation does not mean that these methods perform poorly in these cases. In fact, there is not much variation in the candidate estimators' MSEs, and the regret in Figure~\ref{fig:regret} does not become worse even though the rank correlation decreases.

\end{document}